\documentclass[lettersize,journal]{IEEEtran}
\PassOptionsToPackage{table}{xcolor}
\usepackage{amsmath,amsfonts}
\usepackage{algorithmic}
\usepackage{algorithm}
\usepackage{array}
\usepackage[caption=false,font=normalsize,labelfont=sf,textfont=sf]{subfig}
\usepackage{textcomp}
\usepackage{stfloats}
\usepackage{url}
\usepackage{verbatim}
\usepackage{graphicx}
\usepackage{cite}

\usepackage{hyperref}

\usepackage{url}
\usepackage{graphicx}
\usepackage{multirow}
\usepackage{booktabs}
\usepackage{float}

\newcommand{\pv}[0]{\ensuremath{\boldsymbol{p}} }

\newcommand{\wv}[0]{\ensuremath{\boldsymbol{w}} }
\newcommand{\xv}[0]{\ensuremath{\boldsymbol{x}} }

\newcommand{\thetav}[0]{\ensuremath{\boldsymbol{\theta}} }

\usepackage{float}                  
\usepackage{subfig}              
\usepackage{overpic}   
\usepackage{algorithm}
\usepackage{algorithmic}
\usepackage{tikz}
\usepackage{amsmath}
\usepackage{xcolor}

\begin{document}

\title{LLM Empowered Prototype Learning for Zero and Few-Shot Tasks on
Tabular Data}

\author{Peng Wang, Dongsheng Wang, He Zhao, \textit{Member, IEEE}, Hangting Ye, Dandan Guo*, Yi Chang, \textit{Senior Member, IEEE}
\thanks{
This work was supported by NSFC under Grant 62306125, Grant 2023YFF0905400, and Grant U2341229. (*Corresponding author: Dandan Guo.)

Peng Wang, Hangting Ye, Yi Chang, and Dandan Guo are with the School of Artificial Intelligence, Jilin University, China. Yi Chang is also with the Engineering Research Center of Knowledge-Driven
Human-Machine Intelligence, Ministry of Education, China and the International Center of Future Science, Jilin University, China.  E-mail: pwang23@mails.jlu.edu.cn, yeht2118@mails.jlu.edu.cn, yichang@jlu.edu.cn, guodandan@jlu.edu.cn.

Dongsheng Wang is with the College of Computer Science and Software Engineering, Shenzhen University, China. E-mail: dongshengwang@szu.edu.cn. 

He Zhao is with CSIRO’s Data61 and Monash University, Australia. E-mail: he.zhao@data61.csiro.au.
}}

\markboth{Journal of \LaTeX\ Class Files,~Vol.~14, No.~8, August~2021}%
{Shell \MakeLowercase{\textit{et al.}}: A Sample Article Using IEEEtran.cls for IEEE Journals}


\maketitle

\begin{abstract}


Recent breakthroughs in large language models (LLMs) have opened the door to in-depth investigation of their potential in tabular data modeling. 
However, effectively utilizing advanced LLMs in few-shot and even zero-shot scenarios is still challenging. 
To this end, we propose a novel LLM-based prototype estimation framework for tabular learning.
Our key idea is to query the LLM to generate feature values based example-free prompt, which solely relies on task and feature descriptions. With the feature values generated by LLM, we can build a zero-shot prototype in a training-free manner, which can be further enhanced by fusing few-shot samples, avoiding training a classifier or finetuning the LLMs. Thanks to the example-free prompt and  prototype estimation, ours bypasses the constraints brought by the example-based prompt, providing a scalable and robust framework.
Extensive experiments demonstrate the effectiveness of ours in zero and few-shot tabular learning.





\end{abstract}

\begin{IEEEkeywords}
Tabular data, LLMs, Example-free prompt, Prototype estimation, Zero and few-shot.
\end{IEEEkeywords}
\section{Introduction}

Tabular data, consisting of structured rows and columns, is a critical data format in industries including finance~\cite{arun2016loan,clements2020sequential}, healthcare~\cite{ulmer2020trust,zhou2020pneumonia}, and other fields~\cite{buczak2015survey,guo2017deepfm}.
Few-shot and zero-shot learning~\cite{sanh2021multitask,hegselmann2023tabllm,nam2023stunt}
 have gained increasing attention in tabular data analysis, where models must generalize from limited or even no labeled examples. This is especially critical in real-world domains where data collection and labeling can be expensive, sensitive, or time-consuming. The heterogeneous nature of tabular data and the absence of natural sequential relationships among features make tabular learning challenging. These issues are exacerbated in few-shot settings, where algorithms rely on prior knowledge to learn from limited data. Traditional approaches often fail to perform well under such constraints~\cite{nam2023stunt,hegselmann2023tabllm}. Recently, functional as both knowledge repositories and reasoning engines, LLMs have demonstrated remarkable reasoning abilities and can help solve unseen tasks~\cite{achiam2023gpt,badaro2023transformers,dubey2024llama, ban2025llm}.

To this end, recent research has focused on leveraging LLMs to few-shot tasks.
 Notably, LLMs can quickly adapt to new tasks through task-oriented prompts. This prompt-based approach is not only user-friendly but also significantly enhances the model's ability to transfer pre-trained knowledge to novel and unseen scenarios efficiently. 
In alignment with these principles, recent attempts have designed prompt templates such as ``\textit{\textless Meta-Info\textgreater-\textless Example\textgreater-\textless Query\textgreater}'',  where \textit{\textless Meta-Info\textgreater}, \textit{\textless Example\textgreater} and \textit{\textless Query\textgreater} denote the task and feature descriptions, few-shot examples, and user queries, respectively.
For instance, Summary Boosting conceptualize LLMs as weak learners, treating LLMs as base learners within an ensemble framework~\cite{manikandan2023language}; FeatLLM views LLMs as rule generators and prompts LLMs to directly output decision rules for each class by feeding the few-shot examples, showing an efficient strategy to utilize prior knowledge~\cite{han2024featllm}. Despite their strong performance, these example-driven methods have limitations. Including specific examples in prompts can risk data leakage, leading to suboptimal reasoning. 
LLMs have limited input token lengths, making it difficult to accommodate large sample sizes or high-dimensional features. These issues hinder their scalability and usability in real-world scenarios. Furthermore, the reliance on examples restricts their applicability in zero-shot settings, where no labeled instances are available for prompting.

Recall that prototype learning has emerged as an effective approach for tabular learning~\cite{nam2023stunt,ye2024ptarl}. 
By computing representative embeddings (prototypes) for each class or cluster, models can classify new instances based on their similarity to these learned prototypes, reducing the reliance on large amounts of labeled data. However, conventional prototype learning methods still heavily depend on the availability of high-quality examples, which hinders their applications in zero-shot learning. 
Motivated by recent success in zero-shot learning, where LLMs are shown to reason effectively using only task descriptions~\cite{kojima2022large,wang2023large}, we propose a novel prototype-based approach that leverages LLMs without relying on explicit examples. Specifically, we remove the \textit{\textless Example\textgreater} component from the prompt template and prompt the LLM to generate representative feature values for each class based on task descriptions alone. These generated values are then used to construct class-wise prototypes that capture the semantic characteristics of each class. 
By leveraging the power of LLMs, ours enables prototype learning for zero-shot classification without any example-dependent inference or fine-tuning. In the few-shot setting, we enhance the generated prototypes by fusing them with the embeddings of a few labeled instances from the same class, thus improving representational robustness while preserving scalability. Unlike conventional LLM-based methods that query LLMs for each data sample during inference, our method precomputes class prototypes, allowing for efficient and scalable prediction. In summary, we name our method \textbf{ProtoLLM}, a training-free prototype learning framework with example-free prompt for LLMs that aligns the reasoning abilities of LLMs with the efficiency of prototype learning. 
We compare ProtoLLM with advanced baselines on multiple tabular datasets in zero and few-shot settings and provide detailed ablations from various perspectives, showing its robust and superior performance. It  provides a solid option for future studies on tabular.

ProtoLLM's advantages are highlighted:
\begin{itemize}
    \item \textbf{Training-Free and Example-Free Prompt}. LLMs take the prompt ``\textit{\textless Meta-Info\textgreater-\textless Query\textgreater}'' as input and output the discriminative feature values for each class, resulting in a novel example-free framework for tabular data learning. The class-level prototype then can be estimated without training, which can be used for classification, making ours a training-free method.



    \item \textbf{Feature Value Generation by LLMs}. Unlike traditional data augmentations that employ LLMs to generate all features of a tabular sample simultaneously, we focus on querying LLMs feature by feature. This feature-level generation relieves LLMs from the complex inter-feature relationships, resulting in a meaningful feature discovery. 
    \item \textbf{Zero-shot and few-shot}. Due to the example-free prompt, the generate feature values by LLM can be viewed as the prior
   knowledge in LLM about the task and feature. The few-shot samples can be used to shift the prior feature values derived from LLMs to the target domain through a fusing strategy, achieving an enhanced prototype estimation.  Designed in this way makes ours available and flexible in zero-shot and few-shot settings.
   
   
   
\end{itemize}


\section{Related Work}

\textbf{Few-shot Tabular Data Learning.} 
The development of effective algorithms for few-shot tabular learning has long been a popular research topic, due to their wide variety of applications~\cite{borisov2022deep,kadra2021well,sattarov2023findiff, zhang2024natural}. However, tasks in tabular data are often heterogeneous and interdependent, which makes data collection and annotation challenging~\cite{shan2019crowdsourcing}. These difficulties are further amplified under few-shot settings, where limited labeled examples pose significant challenges for tabular classification and learning. Inspired by the great success across various modalities~\cite{schick2020exploiting,wang2024tuning}, previous works have proposed a number of  frameworks using different techniques, including Bayesian inference, semi-supervised learning, prototype learning and others. For example, TabPFN design a prior to model complex uncertainty of tabular data and show promising performance in small tabular classification tasks~\cite{hollmann2023tabpfn,hollmann2025accurate}. STUNT uses unlabeled data to generate diverse few-shot tasks and construct a prototype classifier~\cite{nam2023stunt}, while some works show that contrastive learning is another option to learn general features~\cite{verma2021towards,shenkar2022anomaly}. 


\textbf{Understanding Tabular Data with LLMs.}
Recent studies have explored integrating tabular data with LLMs across various downstream tasks~\cite{hollmann2024large, seedat2023curated, longbridging, ye2025drl}. In particular, the extensive prior knowledge and language understanding capabilities of LLMs offer a promising foundation for few-shot  and even zero-shot tabular learning scenarios, reducing the reliance on large annotated datasets.  
For instance, LIFT demonstrates that LLMs can be fine-tuned for non-language tasks via natural language interfaces and further extends this framework to support in-context learning and other adaptation methods~\cite{dinh2022lift}. TabLLM serializes tabular data and fine-tunes the T0 model for classification~\cite{hegselmann2023tabllm}.  
P2T proposes a transfer learning framework that utilizes unlabeled source data with LLMs~\cite{nam2024tabular}. They are computationally expensive for using the LLM to make predictions during the test stage. FeatLLM is proposed to avoid fine-tuning LLM and go beyond using the LLMs to predict each test sample~\cite{han2024featllm}. It extracts class-specific rules from few-shot examples with LLM. These rules generate binary features for training a linear classifier, with ensembling through multiple iterations.
However, these above-mentioned methods design the example-based prompt for LLMs. They will face: (1) scalability issues with large samples, features, or the number of classes due to the token length constraint in LLMs; (2) potential privacy leakage for exposing examples to LLMs. (3) poor performance or even inapplicability in zero-shot scenarios.


Ours utilizes LLMs for tabular zero and few-shot learning. The key difference is that we use LLM to estimate the prototype. We design an example-free prompt for LLMs to generate feature values and then construct the training-free prototype for classification. Notably, we do not need to fine-tune LLMs, and we can predict the test samples based on the prototype instead of LLMs, thereby incurring a lower inference cost. Benefiting from the example-free but efficient prompt, ours still has desired performance in zero-shot settings.








\section{Methods} 
\begin{figure*}[!htbp]
    \centering
    \includegraphics[width=0.8\linewidth]{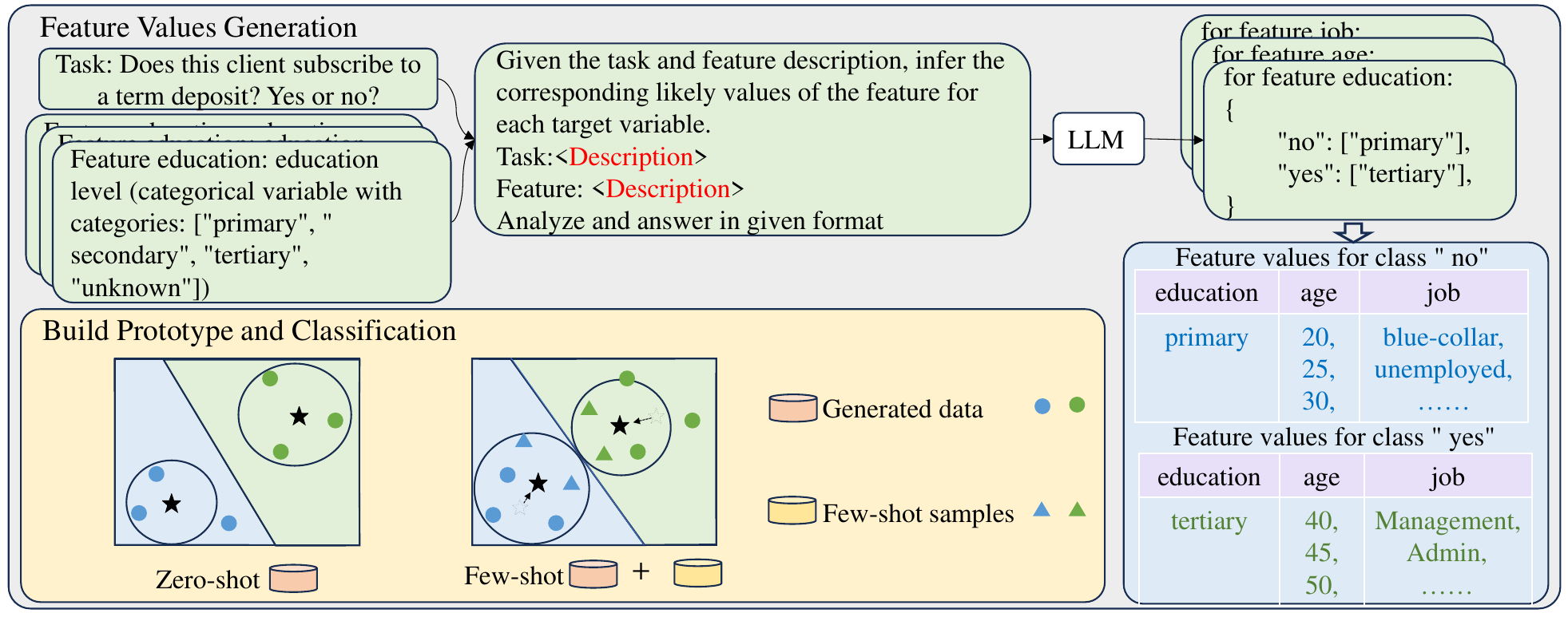}
        \caption{\small{Overview of our ProtoLLM. The upper part shows the example-free feature value generation, which predicts the potential values of feature for each class by LLM, where the prompt depends on the description about task and feature. The lower part builds the prototypes with the generated and few-shot samples, and introduces a training-free classification framework.}
    }
    \label{fig:Generate-data-w-LLM}
\end{figure*}

This work introduces ProtoLLM, a novel training-free and example-free framework for integrating LLMs into tabular few-shot classification, whose overview is shown in Fig.~\ref{fig:Generate-data-w-LLM}. The core idea behind ProtoLLM is to generate discriminative feature values by querying LLMs with example-free prompts, as described in Sec.~\ref{sec:oracle feature generation}. 
We then design a training-free method to build prototypes and classify new samples in Sec.~\ref{sec:classification}. 




\textbf{Problem Formulation.} Denote the tabular dataset as $ \mathcal{S} = \{ (\xv_n, y_n) \}_{n=1}^N $, where the number of samples $N$ is usually small and $N=0$ means the zero-shot setting. Each sample $ \xv_n $ consists of $ D $ features in total. $ x_n^d $ is the $ d $-th feature of $\xv_n$, which can be either a numerical or a categorical feature. Specifically, if $ x_n^d $ is a numerical feature, then $ x_n^d \in \mathbb{R} $ represents a scalar value. If $ x_n^d $ is  categorical, we represent it as the one-hot encoded vector with $1$ denoting the corresponding category. Moreover, $ y_n \in \{1, \ldots, C\} $ means the label of sample $\xv_n$, with $ C $ being the total number of classes. Denote $\mathcal{F} =\{f_{\text{task}}, f_{\text{feat}}^{1:D}\}$ as the set of descriptive information for the dataset, which are usually available in tabular datasets. Specifically, $f_{\text{task}}$ is the information related to the task and $f_{\text{feat}}^d$ is the descriptions about the $d$-th feature. For example, if the $d$-th feature in the Adult dataset is ``education'', $f_{\text{feat}}^d$ includes a description of the feature, such as ``education level'',  and its possible values, e.g., primary, secondary, tertiary, unknown. 


\subsection{Feature Value and Importance Generation by LLM}\label{sec:oracle feature generation}
\begin{figure*}[!t]
\centering
    \subfloat[prompt ]{\includegraphics[width=0.99\columnwidth]{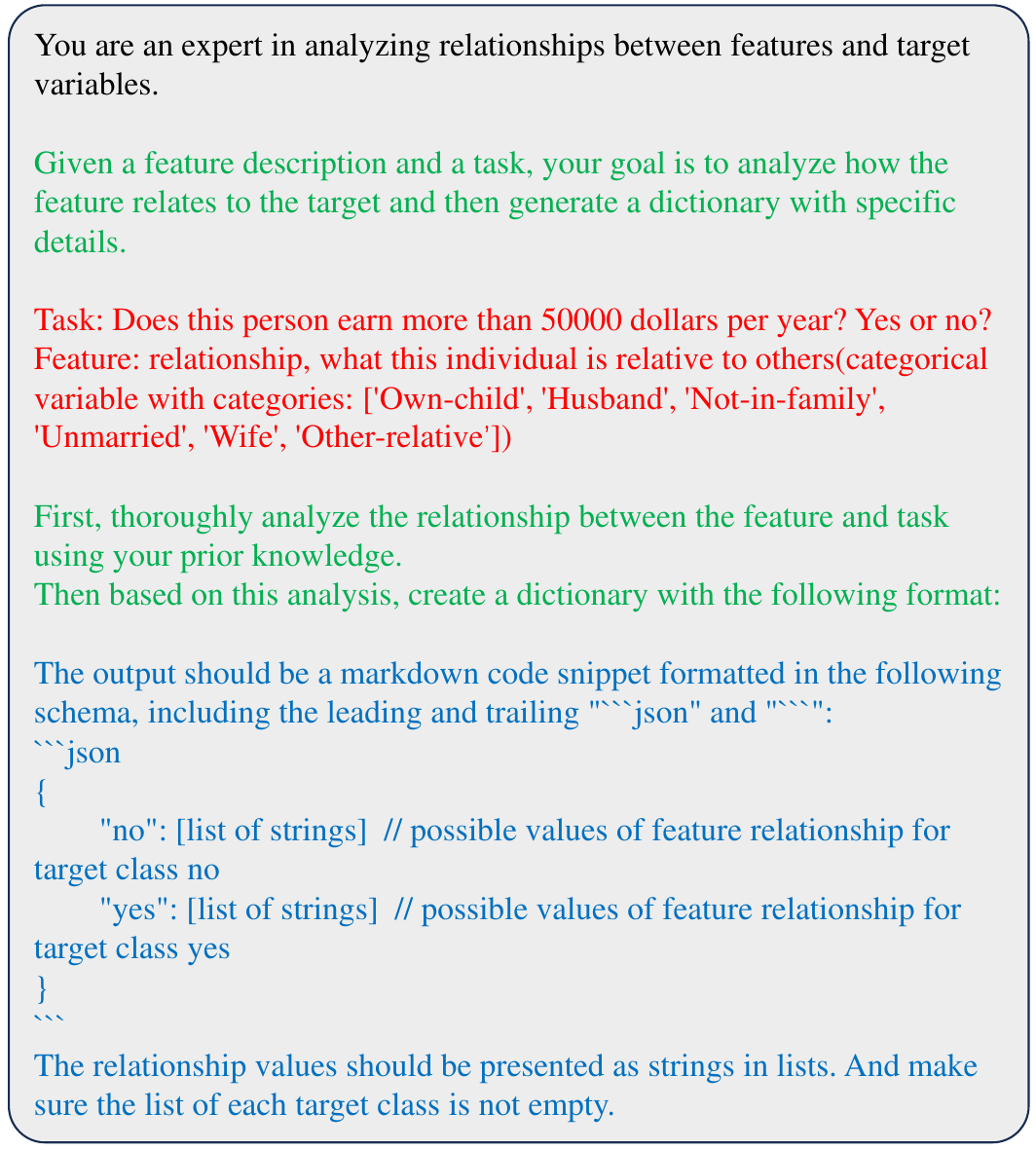}
    \label{fig:prompt}
    }
    \quad
    \subfloat[answer]{\includegraphics[width=0.99\columnwidth]{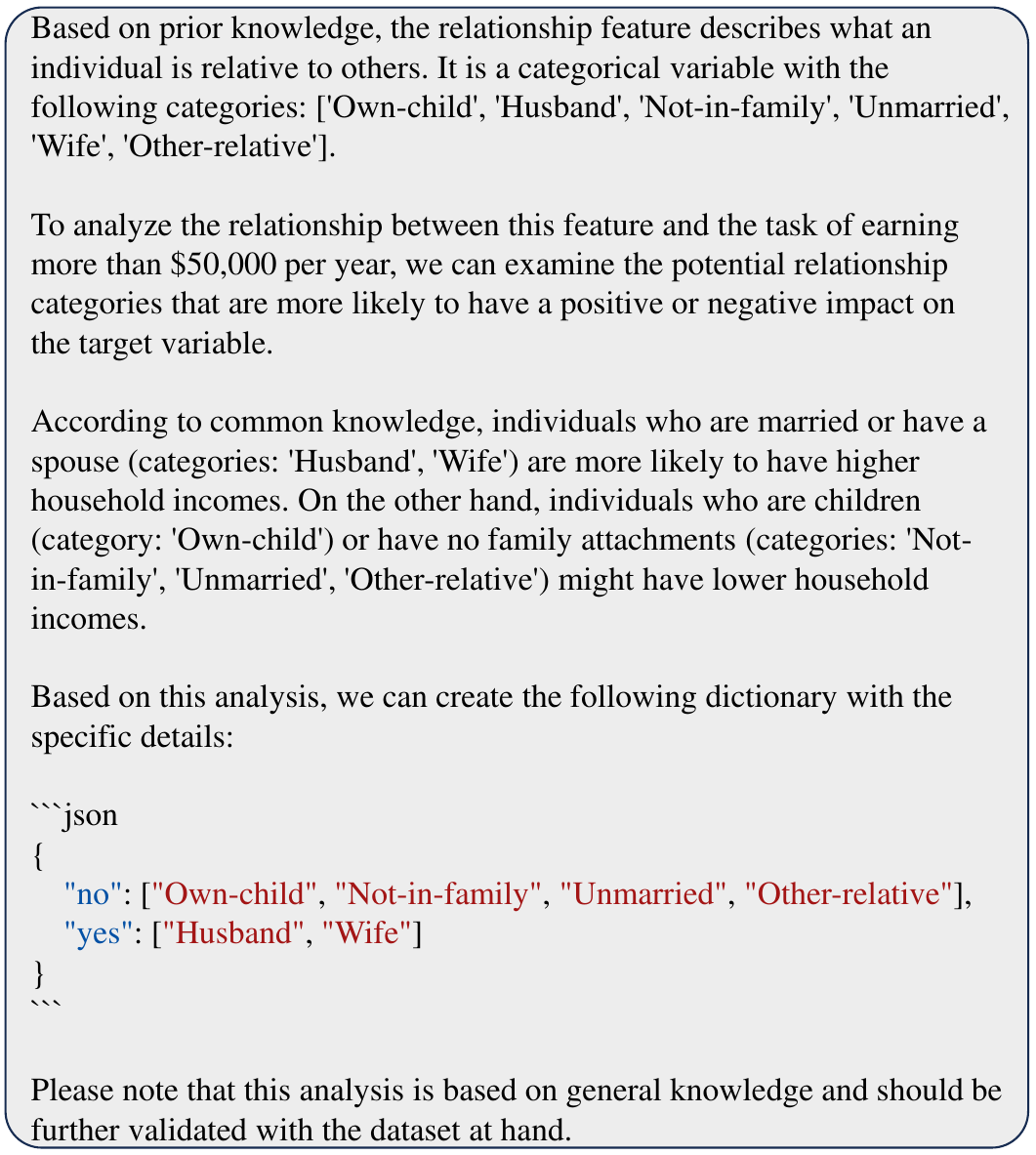}
    \label{fig:answer}
    }
    \caption{{(a) Our designed prompt for generating feature values about the  ``relationship'' attribute in Adult dataset. Here, black and red words mean the classic sentence in prompt and descriptions about task and feature, i.e., \textit{\textless Meta-Info\textgreater}. Besides, green words are the reasoning instruction and blue words denote the expected response format, which construct the \textit{\textless Query\textgreater}. (b) The corresponding response generated by GPT-3.5 for the prompt, where ``no'' and ``yes'' in blue color denote the target class (whether this person earns more than 50K dollars per year?). Besides, the red words mean the feature values generated by LLMs.}}
    \label{fig:prompt-answer}
\end{figure*}

To leverage the prior knowledge of LLMs in our problem, as illustrated in Fig.~\ref{fig:prompt-answer}, we carefully design an example-free prompt by proposing a novel way. Specifically, for $d$-th feature, we denote the prompt as $P(f_{\text{task}},f_{\text{feat}}^d)=[\textit{\textless Meta-Info\textgreater-\textless Query\textgreater}]$, where we shorten it to $P^{\text{value}}_d$ for convenience.
The prompt is started with ``you are an expert in ...'', a classic and shared sentence in prompt engineering. Then, we design ``\textit{\textless Meta-Info\textgreater}'' by introducing the information about task and the $d$-th feature. It is used to provide basic information into LLMs. Besides, ``\textit{\textless Query\textgreater}'' is constructed with a reasoning instruction followed by a requirement of output formation. For the $d$-th feature, we query LLMs with the prompt template $P^{\text{value}}_d$ and expect that LLMs output correct feature values for each class. We describe this process in more detail below.

\textbf{Task and Feature Information in \textit{\textless Meta-Info\textgreater}}. For LLMs, it's crucial to provide a clear task description and detailed feature information. The task description explains the objective, scope, and expected outcomes. As shown in Fig.~\ref{fig:prompt-answer}, we summarize the \textit{\textless Meta-Info\textgreater} in red words. For the Adult dataset, the task is “\textit{Does this person earn more than 50000 dollars per year? Yes or no?}”. Each feature description outlines the input variables used in prediction, clarifying their roles. For instance, “relationship” refers to the individual's family role, such as spouse or child, which can impact income potential. The description is “\textit{What this individual is relative to others}”. Designed in this way, the meta-information (task and feature descriptions) allows LLMs to understand the current task and leverage their prior knowledge for generating representative feature values. 
 
\textbf{Reasoning Instruction \& Response Format in \textit{\textless Query\textgreater}}. The objective of the prompt 
is to guide the LLM in generating possible values for each target class based on a given feature. 
Motivated by the recent chain-of-thought (CoT) tricks~\cite{wei2022chain,lyu2023faithful}, 
we construct the reasoning instruction with two steps, i.e., green words in Fig.~\ref{fig:prompt-answer}. 
First, we ask LLMs to analyze the potential causality of the task and feature based on 
the provided information in ``\textit{\textless Meta-Info\textgreater}''. This allows LLMs to 
mobilize their pre-trained knowledge about the question. Given the output analysis, LLMs are 
encouraged to infer the possible feature values for each class, where we devise the response 
format with blue words, as shown in Fig.~\ref{fig:prompt-answer}. For numerical features, since their range information is often difficult to obtain or may be unavailable in practice, we additionally include the following prompt to help infer the range of numerical features: \textit{``numeric variable, you should use your prior knowledge to determine the appropriate ranges of values.''}. Examples are provided in Appendix~\ref{appendix: Example of prompt}. 
Once we construct the prompt $P^{\text{value}}_{d}$ for $d$-th feature, we can query LLM with $P^{\text{value}}_{d}$, where we denote LLM 's output as  $\text{LLM}(P^{\text{value}}_{d})$. Specifically, let $z_{c,d}$ denote the generated 
value for $d$-th feature in class $c$:
\begin{equation}\label{generated_feature}
\left\{
\begin{aligned}
z_{c,d} &= \text{One-hot}(\text{LLM}(P^{\text{value}}_{d})[c]), \text{ categorical} \\
z_{c,d} &= \text{LLM}(P^{\text{value}}_{d})[c], \text{numerical}
\end{aligned}
\right.
\end{equation}
where categorical denotes categorical feature, numerical denotes numerical feature and $\text{LLM}(\cdot)[c]$ 
denotes the output values of LLMs for $c$-th class. For  numerical features, we directly use the output values of LLMs.  For categorical 
features, we apply the $\text{One-hot}(\cdot)$ function to convert the categorical 
features into vectors, where LLMs output $m$ feature values, $z$ is 
obtained as:$z=\frac{1}{m}\sum_m z^{m}$. Considering the robustness, we can query LLMs $K$ times 
for each feature independently, resulting in a set of feature values: $\mathcal{Z}_{c,d} = \{z_{c,d}^k\}_{k=1}^K$, 
where our approach requires $K\times D$ queries.



\begin{figure*}[!t]
\centering
    \subfloat[prompt ]{\includegraphics[width=0.99\columnwidth]{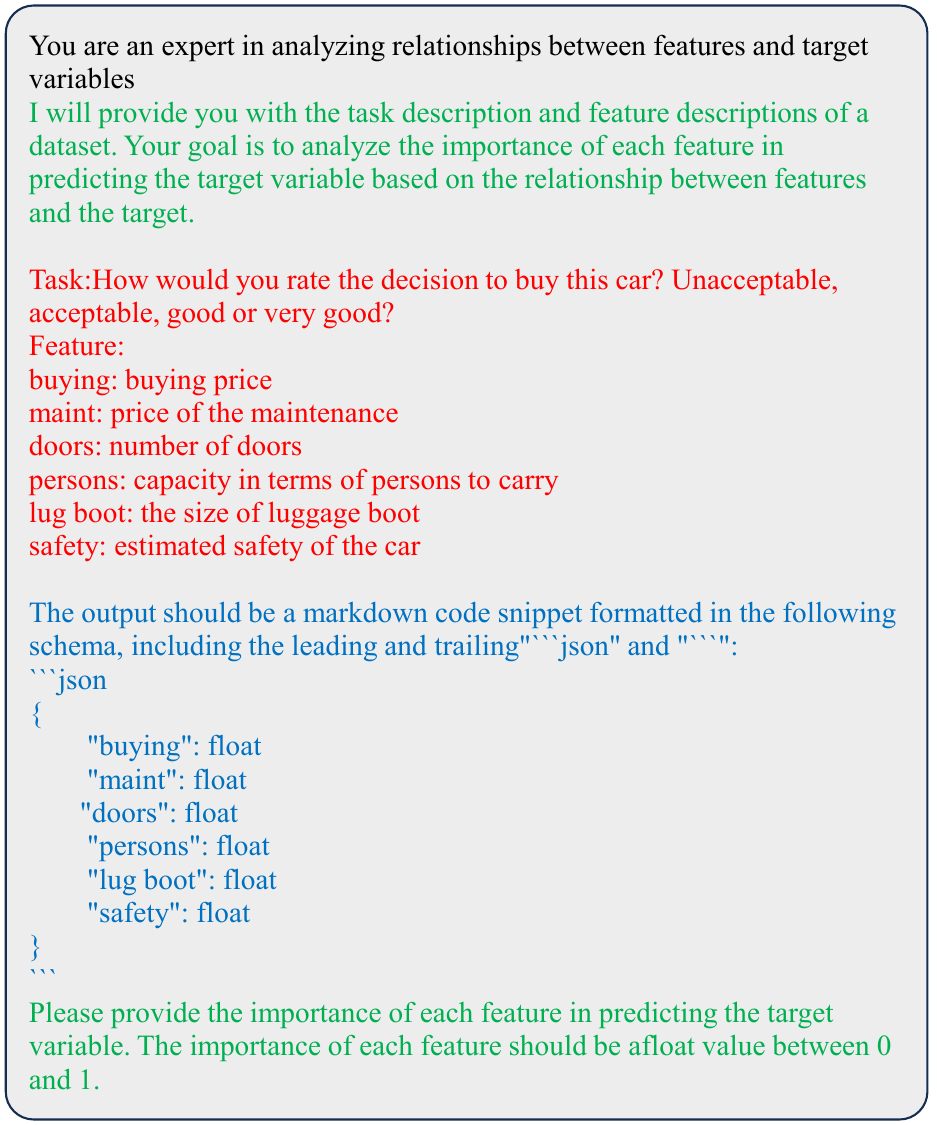}
    \label{fig:weight-prompt}
    }
    \quad
    \subfloat[answer]{\includegraphics[width=0.99\columnwidth]{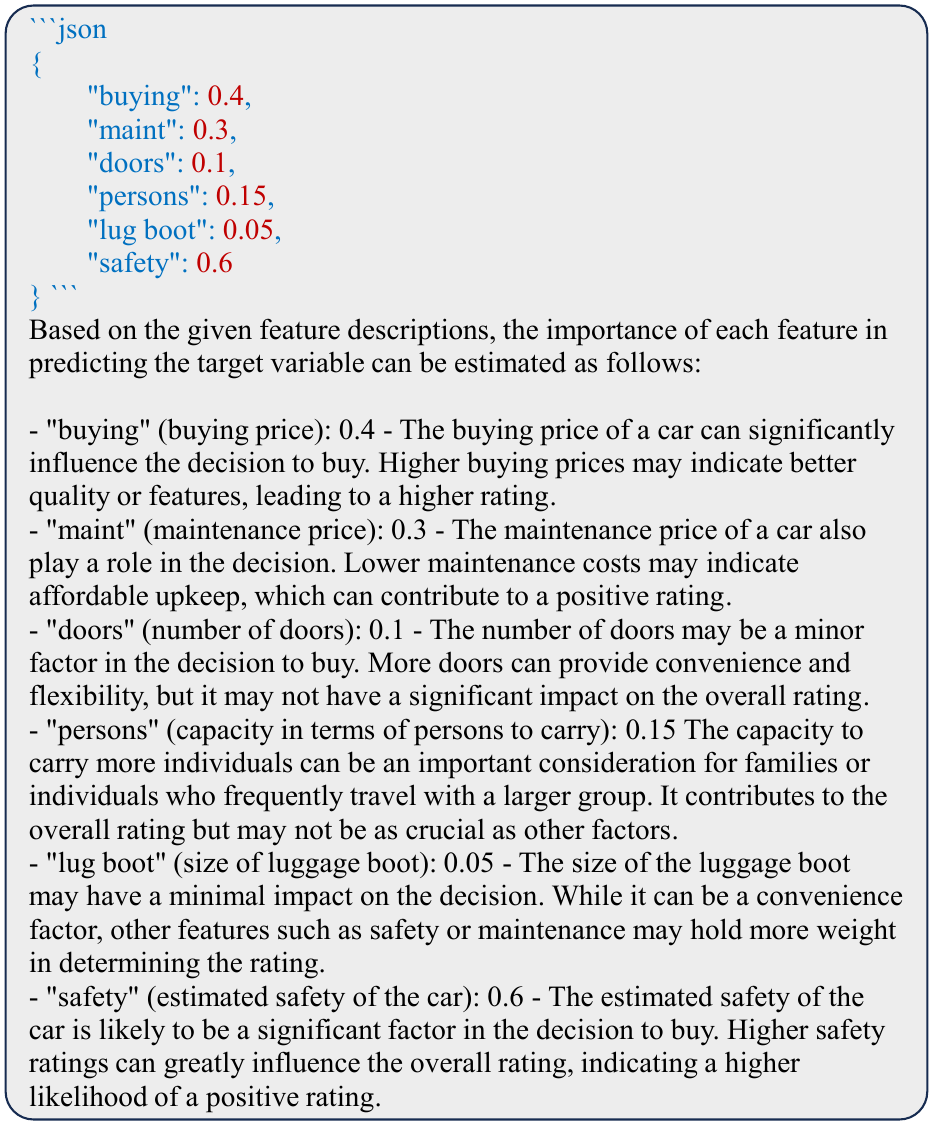}
    \label{fig:weight-answer}
    }
    \caption{(a) The designed prompt used to query feature importance within the dataset. We denote the descriptions of each feature from the dataset along with the task definition as  \textit{\textless Meta-Info\textgreater}, to guide the LLM in analyzing the relative importance of each feature  i.e., red words. Besides, green words are the reasoning instruction and blue words denote the expected response format, which construct the \textit{\textless Weight Query\textgreater}.
   (b) The corresponding response generated by GPT-3.5 for the prompt, which includes a JSON-formatted dictionary indicating the importance scores of each feature, along with explanations for the reasoning behind these assessments.
}

    \label{fig:weight-prompt-answer}
\end{figure*}

\textbf{Feature Importance Generation by LLM.} Considering the feature-level query may overlooks the relationships 
between features, we further prompt LLMs to output the importance of each feature in tabular data, 
identifying which features contribute more significantly to the overall prediction. As shown in  Fig.~\ref{fig:weight-prompt}, we provide a example of querying feature importance. For this querying process, both the task and all relevant feature descriptions are provided to the
LLMs, denoted as \textit{\textless Weight Meta-Info\textgreater} (red words), to ensure a comprehensive understanding of the context and specific characteristics of the data. Similar to feature value generation, we also design a simple reasoning instruction (green words)  and expected response format  (blue words), denoted as \textit{\textless Weight Query\textgreater}. Therefore, we denote the prompt for feature importance as $P^{\text{weight}} = P(f_{\text{task}}, f_{\text{feat}}^{1:D}) = [\textit{\textless Weight Meta-Info\textgreater-\textless Weight Query\textgreater}]$. Specifically,  we querying LLM with the designed $P^{\text{weight}}$,  and express the output of LLMs as follows:
\begin{equation}
\begin{aligned}
[w_1, \dots, w_d, \dots, w_D] = \text{LLM}(P^{\text{weight}}), 
\end{aligned}\label{weight_output}
\end{equation}
then we normalize the output weights to ensure they sum to one $\hat{w}_d \!=\! {w_d}/{\sum_{d'=1}^D w_{d'}}$ and let $\hat{\wv}\!=\!\{w_d\}_{d=1}^{D}$  as the normalized weight vector of features for the given task. We only need to query LLMs one time, which can output reasonable feature weights in most cases. Fig.~\ref{fig:weight-answer} is the corresponding answer for prompt $P^{\text{weight}}$, where LLMs first analyze the relationships between features and the target class, then provide a dictionary that delineates the associated feature weights. The generated feature weights could further enhance ProtoLLM.


\subsection{Prototype Construction and Inference}\label{sec:classification}
\textbf{Prototype Construction}. For \textbf{zero-shot} tasks, given the representative feature values, how to utilize them for tabular zero-shot and few-shot classification tasks is  a key problem. Different from most of existing methods that use the augmented samples by LLM to train a classifier, we introduce a training-free method. For zero-shot setting, we denotes $\Theta_c$ as the prototype  of class $c$, and adopt an average strategy on $\mathcal{Z}$:
\begin{equation}\label{zeroshot_proto}
\begin{aligned}
    \Theta_c &= [\hat{w}_1\thetav_{c,1},...,\hat{w}_d\thetav_{c,d},...,\hat{w}_D\thetav_{c,D}], \\
    \thetav_{c,d} &= \frac{1}{K} \sum_{k=1}^K z_{c,d}^k,
\end{aligned}
\end{equation}
where we re-weight each feature and concatenate reweighted features to  build the final prototype $\Theta_c$. Note that this prototype is inferred solely from LLMs via our example-free prompt, and it thus implicitly encodes common knowledge of LLMs for the target task. 
The prototype provides meaningful priors for prototype learning. 

For \textbf{few-shot} tasks, the prototype in Eq.~\ref{zeroshot_proto} can be simply enhanced if the few-shot samples are given (we still use $\thetav_{c,d}$ for simply):
\begin{equation}\label{fewshot_proto}
    \thetav_{c,d} = \frac{1}{K + |\mathcal{S}_c|} \left( \sum_{k=1}^K z_{c,d}^k + \sum_{\xv_n \in \mathcal{S}_c} x_n^d \right),
\end{equation}
where $\mathcal{S}_c$ is the subset of $\mathcal{S}$ containing samples with label $c$. The first prior term focuses on general knowledge from LLMs, which presents the common sense of the given task. The second term can be explained as the data likelihood. It contains the domain information encoded in the input samples.
Eq.~\ref{fewshot_proto} receives information from two different domains and combines them via an average operation.  This simple yet efficient fusing strategy helps the pre-trained knowledge transfer to the target distribution, improving the prototype  in  few-shot setting.

\textbf{Inference.}
Once the prototype is calculated, one can predict the label $y$ for a test sample $\xv$:
\begin{equation}\label{predict}
    p(y\!=\!c|\xv) \!=\! \frac{\text{exp}(-\text{Dist}(\Theta_c, \boldsymbol{\hat{w}}\xv))}{\sum_{c'=1}^C \text{exp}(-\text{Dist}(\Theta_{c'}, \boldsymbol{\hat{w}}\xv))},
\end{equation}
where 
$\text{Dist}(\cdot,\cdot)$ measures the distance between the prototypes and feature-weighted sample $\boldsymbol{\hat{w}}\xv$, which is specified as the Euclidean distance by default. Notably, Eqs.~\ref{zeroshot_proto}-~\ref{predict} are calculated without any learnable variables, resulting in a training-free framework. Moreover, once the prototypes are learned, {no LLM queries need to be made at inference time}.

\subsection{ProtoLLM Workflow and Further Analysis}
Our workflow is summarized in Algorithm~\ref{algo1}. At step (1), we design an example-free prompt for LLMs to  build a set of feature values $\mathcal{Z}$; at step (2), we further introduce a prompt for LLM to generate the feature importance; at step (3), we construct a training-free prototype for zero-shot task and few-shot task, respectively; at step (4), we implement the prediction for the test sample.

\begin{algorithm}[h]
\caption{Workflow of our proposed ProtoLLM.}
\label{algo1}
\begin{algorithmic}[1]

\STATE \textbf{Require:} Dataset $\mathcal{S}\!=\!\{  ({\xv}_n,y_n) \}_{n=1}^N$ for few-shot and $\mathcal{S}\!=\!\emptyset$ for zero-shot, descriptive information $\mathcal{F}=\{f_{\text{task}},f_{\text{feat}}^{1:D}\}$ about dataset $\mathcal{S}$, test sample $\xv$, LLM;
\STATE \textbf{Output:} Predicted probability vector $\pv$ for test sample; 
\STATE \textbf{Step 1:} {Generate $\mathcal{Z}$ by LLM with our designed example-free prompt and initial $\mathcal{Z}=\emptyset$ as:}
\FOR{$d=1$ to $D$}
\STATE Design prompt as $P(f_{\text{task}},f_{\text{feat}}^d)=[\textit{\textless Meta-Info\textgreater-\textless Query\textgreater}]$, abbreviated as  $P_d$;
\FOR{$k=1$ to $K$}
\STATE Prompt LLM with $P_d$ and output $z_{c,d}^k$ with Eq. \ref{generated_feature}; set $\mathcal{Z}=[\mathcal{Z},z_{c,d}^k]$; 
 \ENDFOR
 \ENDFOR
 
\STATE \textbf{Step 2:} {Generate global feature weights by LLM:}
\STATE Design prompt as $P^{\text{weight}} = P(f_{\text{task}}, f_{\text{feat}}^{1:D}) = [\textit{\textless Weight Meta-Info\textgreater-\textless Weight Query\textgreater}]$;
\STATE Prompt LLM with $P^{\text{weight}}$ and obtain normalized feature weight vector $\boldsymbol{\hat{w}} = [\hat{w}_1, \hat{w}_2, \dots, \hat{w}_D]$ with Eq.\ref{weight_output};

\STATE \textbf{Step 3:} {Build prototypes $\Theta_{1:C}$ with $\mathcal{Z}$ and  $\mathcal{S}$ in Eq. \ref{zeroshot_proto} for zero-shot or Eq. \ref{fewshot_proto} for few-shot;}
\STATE \textbf{Step 4:} {Compute the predicted probability vector $\pv$ for test sample $\xv$ with Eq. \ref{predict};}
\STATE \textbf{return} $\pv$ \hfill $\triangleright$ Return predicted probabilities
\end{algorithmic}
\end{algorithm}

Here, we analyze our method in more depth from several critical perspectives. (1) \textbf{Example-free prompt for enhanced generalization.} ProtoLLM generates feature values using an example-free prompt, relying solely on task and feature descriptions. This design mitigates the limitations posed by example-based prompting, especially in few-shot scenarios. When the sample size is extremely small, selected examples are often not representative of the overall data distribution and may introduce bias. In contrast, by eliminating dependency on specific examples, our method enables LLMs to rely more heavily on their own internalized prior knowledge. (2) \textbf{Decomposition into sub-problems for better reasoning.} ProtoLLM focuses on a single feature at each query time, decomposing the original multi-feature reasoning task into $D$ more tractable sub-problems. Although these sub-problems are not completely independent, solving them individually reduces the complexity of each generation step, making the task more manageable for LLMs. This design simplifies the reasoning process and improves generation stability, particularly in tabular data where feature heterogeneity and complex inter-feature relationships pose significant challenges for joint reasoning. (3) \textbf{Prototype construction via LLM prior and domain adaptation.} ProtoLLM calculates prototypes by explicitly combining two complementary sources of information: prior knowledge from the LLM and empirical statistics from few-shot samples. This fusion enables the model to capture both global, concept-level information and localized, task-specific adaptation. Notably, the entire framework is training-free and requires no gradient updates, making it easy to deploy, adaptable to new tasks, and friendly to environments with limited computational resources. (4) \textbf{Additional example for regression task.} Mainly targeting zero-shot and very few-shot classification for tabular data,  we also exploit its potential on tabular regression tasks shown in the Appendix~\ref{appendix: Other tasks}.

\section{Experiments}

\textbf{Datasets.}
Following  FeatLLM~\cite{han2024featllm}, we conduct few-shot tabular data classification across 10 datasets, including binary or multi-class classification tasks. Besides, we also conduct zero-shot experiments. Specifically, we use Adult~\cite{kohavi1996scaling}, Bank~\cite{moro2014data}, Blood~\cite{yeh2009knowledge}, Car~\cite{kadra2021well}, Credit-g~\cite{kadra2021well}, Diabetes~\cite{smith1988using}, Heart~\cite{detrano1989international}, Myocardial~\cite{golovenkin2020trajectories}, and {two other datasets including Cultivars~\cite{cultivars2023dataset} and NHANES, which were released recently and not included in the LLMs pre-training stage~\cite{han2024featllm}}. These datasets cover fields such as financial, medical, and recommendation, varying in size and complexity. Each dataset contains the corresponding name and description about features. 
The properties about all of the datasets are shown in Tab.~\ref{tab:dataset info} and a summary of the key characteristics and classification objectives of each dataset is provided below:

\begin{table*}[htbp]
    \centering
    \caption{Properties about all of the datasets.}
    \resizebox{0.8\linewidth}{!}{
    \begin{tabular}{ccccccccccc}
    \toprule
            & Adult & Bank  & Blood & Car   & Credit-g & Cultivars & Diabetes & Heart & Myocardial & NHANES \\ 
    \midrule
        Objects & 48842 &45211&748&1728&1000&320&768&918&1700&6287\\

        Numerical & 7 & 8 & 4 & 1 & 8 & 7 & 8 & 7 & 17 & 7 \\

        Categorical & 7 &8 &0 &5 &12 &3 &0 &4&94 & 1 \\

        Class &2 &2 &2 &4 &2 &2 &2 &2 &2 &2 \\
    \bottomrule
        \end{tabular}
    }
    
    \label{tab:dataset info}
\end{table*}

\begin{itemize}
    \item The \textbf{Adult}\footnote{\small{\href{https://archive.ics.uci.edu/dataset/2/adult}{archive.ics.uci.edu/dataset/2}}} dataset is used to determine whether an individual earns more than \$50,000 annually, based on demographic and employment features.
    \item The \textbf{Bank}\footnote{\small{\href{https://archive.ics.uci.edu/dataset/222/bank+marketing}{archive.ics.uci.edu/dataset/222}}} dataset predicts whether a customer will subscribe to a term deposit, utilizing personal and socio-economic factors.
    \item The \textbf{Blood}\footnote{\small{\href{https://archive.ics.uci.edu/dataset/176/blood+transfusion+service+center}{archive.ics.uci.edu/dataset/176}}} dataset is designed to forecast whether a person will donate blood, given past donation records.
    \item The \textbf{Car}\footnote{\small{\href{https://archive.ics.uci.edu/dataset/19/car+evaluation}{archive.ics.uci.edu/dataset/19}}} dataset classifies the acceptability of a car based on attributes like buying price, maintenance cost, and safety features.
    \item The \textbf{Credit-g}\footnote{\small{\href{https://archive.ics.uci.edu/dataset/144/statlog+german+credit+data}{archive.ics.uci.edu/dataset/144}}} dataset addresses the classification of individuals as good or bad credit risks, using personal and financial attributes.
    \item The \textbf{Cultivars}\footnote{\small{\href{https://archive.ics.uci.edu/dataset/913/forty+soybean+cultivars+from+subsequent+harvests}{archive.ics.uci.edu/dataset/913}}} dataset assesses the growth and yield of forty soybean cultivars under varying conditions, with features such as plant height, number of stems, and grain yield.
    \item The \textbf{Diabetes}\footnote{\small{\href{https://www.kaggle.com/datasets/uciml/pima-indians-diabetes-database}{kaggle.com/datasets/uciml/pima-indians-diabetes-database}}} dataset focuses on predicting the likelihood of diabetes in a person, relying on medical metrics such as glucose levels and BMI.
    \item The \textbf{Heart}\footnote{\small{\href{https://www.kaggle.com/datasets/fedesoriano/heart-failure-prediction}{kaggle.com/datasets/fedesoriano/heart-failure-prediction}}} dataset identifies whether a patient is at risk of heart disease, considering factors like age, cholesterol levels, and blood pressure.
    \item The \textbf{Myocardial}\footnote{\small{\href{https://archive.ics.uci.edu/dataset/579/myocardial+infarction+complications}{archive.ics.uci.edu/dataset/579}}} dataset is used to predict the outcomes of patients following a myocardial infarction, based on clinical data such as heart rate and blood pressure.
    \item  The \textbf{NHANES}\footnote{\small{\href{https://archive.ics.uci.edu/dataset/887/national+health+and+nutrition+health+survey+2013-2014+(nhanes)+age+prediction+subset}{archive.ics.uci.edu/dataset/887}}} dataset is derived from the National Health and Nutrition Examination Survey, focusing on predicting respondents' age using features such as physiological measurements, lifestyle factors, and biochemical markers.
\end{itemize}





    

\textbf{Baselines and Details.} 
For few-shot settings, we compare ProtoLLM against two types of baselines 
\textit{(1)} Traditional machine learning methods, including Multilayer Perceptron (MLP), Logistic Regression (LogReg), XGBoost(XGB), K-Nearest Neighbors (KNN), and popular few-shot learning methods, including STUNT~\cite{nam2023stunt} and TabPFN\_v2~\cite{hollmann2025accurate}
\textit{(2)} LLM-based framework. This includes methods such as In-context Learning~\cite{nam2023semi}, TABLET~\cite{slack2023tablet},  TabLLM~\cite{hegselmann2023tabllm}, FeatLLM~\cite{han2024featllm}, LIFT~\cite{dinh2022lift} and P2T~\cite{nam2024tabular}. For zero-shot setting, except for P2T and TabLLM,  we also use DSPy~\cite{khattab2023dspy}, which offer an automated solution for enhanced prediction. 
 We set query times $K$ as 10, and use Euclidean distance in Eq.~\ref{predict} in all experiments. Following FeatLLM, we use $N$ labeled samples for the $N$-shot setting, and select \textit{gpt-3.5-turbo-0613} as our base LLM by default. The LLM version is same with our baselines except for TabLLM, which adopts T0 as the backbone for the need of fine-tuning LLM.
 The results of STUNT, In-context learning, TABLET, TabLLM, and FeatLLM are derived from ~\cite{han2024featllm}. We run ProtoLLM 15 times, only with different seeds, and report the average area under the receiver operating characteristic curve (AUC). More details can be found in Appendix~\ref{appendix: baseline details}.


\begin{figure*}
    \centering
    \includegraphics[width=\linewidth]{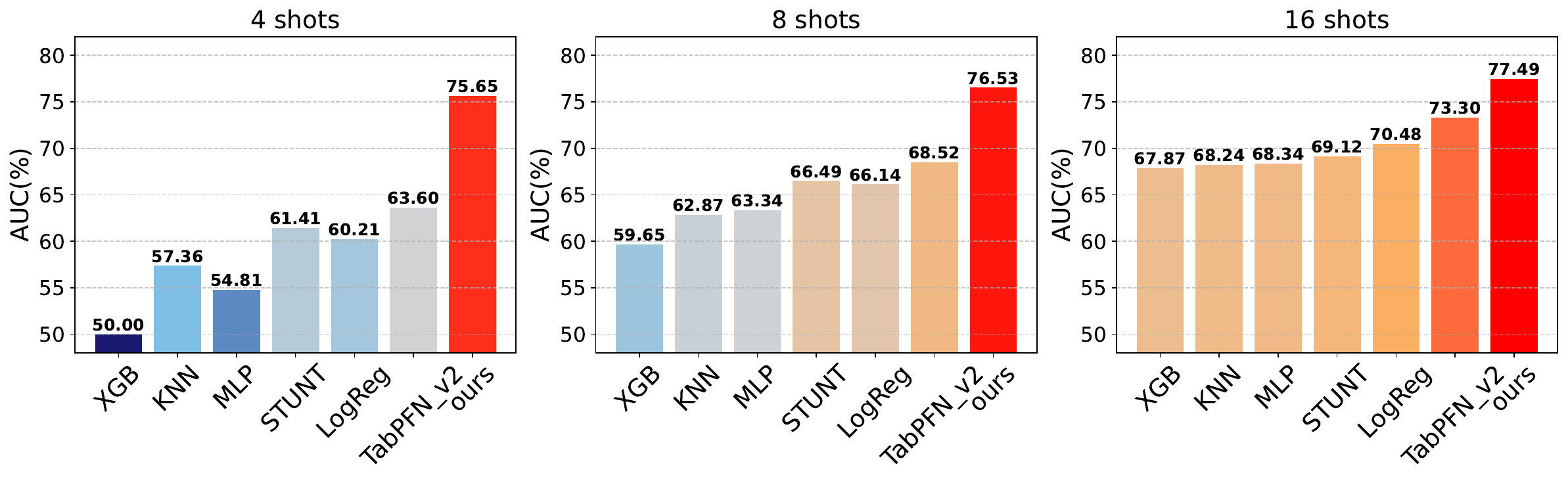}
    \caption{Comparison of the average AUC across datasets  between non-LLM methods and ours.}
    \label{fig:non-llm-results}
\end{figure*}
\subsection{Zero and few-shot Classification}


\textbf{Zero-shot Classification.} As listed in  in Tab.~\ref{tab:zero-shot}, we report the zero-shot results of different methods, where we also query GPT 3.5 for TabLLM to ensure a fair comparison with others. Zero-shot classification remains challenging for tabular datasets. Existing approaches, including P2T, TabLLM, DSPy,  require querying the LLM for each test instance, which incurs significant inference costs and limited to token length in LLM. Besides, P2T needs to utilize the unlabeled data when performing zero-shot task. In contrast, our method supports zero-shot scenarios through example-free prompt and prototype estimation, eliminating the need for LLM inference at test time. As shown in our experiments, it consistently outperforms existing methods across most datasets.

\begin{table*}[!t]
  \centering
  \caption{Performance comparison across 10 datasets in zero-shot scenarios. ``*'': ours without using feature weights in \eqref{weight_output}, ``N/A'': not available due to limited token length. ``Data'': additional data, ``E2E'': end-to-end inference with LLM during testing.}
  \resizebox{\textwidth}{!}{
\begin{tabular}{c|cc|cccccccccccc}
\toprule
\multirow{2}[4]{*}{Method} & \multicolumn{2}{c|}{Requirement} & \multicolumn{11}{c}{Datasets}                                                         &  \\
\cmidrule{2-15}      & Data & E2E   & Adult & Bank  & Blood & Car   & Credit-g & Cultivars & Diabetes & Heart & Myocardial & NHANES & Average (w/o Myocardial) & Average Rank \\
\midrule
TabLLM(T0) & -    & \checkmark   & 80.41 & 62.82 & 53.35 & 49.35 & 49.95 & 50.22 & 59.07 & 47.55 & N/A   & 97.05 & 61.09 & 3.89 \\
TabLLM(GPT-3.5) & -    & \checkmark   & \textbf{87.24} & \underline{69.54} & 61.51 & \underline{75.8} & 45.88 & 49.3  & \underline{79.92} & 64.62 & N/A   & 94.53 & \underline{69.82} & 2.89 \\
DSPy  & -    & \checkmark   & 54.84 & 53.14 & 50.33 & 60.29 & 45.6  & \underline{51.6} & 64.87 & 47.9  & N/A   & 75.07 & 55.96 & 4.22 \\
P2T   & \checkmark   & \checkmark   & 80.73 & 61.12 & \underline{61.9} & 49.89 & \underline{52.14} & 49.71 & 71.06 & \textbf{72.61} & N/A   & \textbf{99.69} & 66.54 & \underline{2.67} \\
\rowcolor[rgb]{ .906,  .902,  .902} Ours   & -    & -    & \underline{83.72} & \textbf{82.93} & \textbf{75.31} & \textbf{77.76} & \textbf{59.64} & \textbf{59.12} & \textbf{80.32} & \underline{68.51} & \textbf{62.52*} & \underline{97.48} & \textbf{76.09} & \textbf{1.3} \\
\bottomrule
\end{tabular}%
    }
  \label{tab:zero-shot}%
\end{table*}%

\textbf{Few-shot Classification.} 
We further report the performance of LLM-based few-shot methods in Tab.~\ref{tab: sub_main_results}. The proposed method yields either the best or highly competitive results across all datasets, achieving the top average AUC and ranking. Besides, when meeting the larger shots or high-dimensional features, many methods with example-based prompts are not available due to the token length limitations in LLM. ProtoLLM is still available and effective by ignoring the feature importance in Eq.~\eqref{weight_output} for dataset with high-dimensional features. We attribute this to the effectiveness of our example-free prototype generation framework. On the one hand, the feature value generation prompts distill reliable prior knowledge from LLMs, providing more accurate feature values for downstream tasks. On the other hand, our simple yet effective prototype fusion strategy efficiently integrates the LLM’s knowledge with few-shot samples, resulting in higher-quality prototypes and improved classification performance. The results with variance are provided in Tab.~\ref{tab: main_results} 
 in Appendix~\ref{appendix:detail_results}. 
We also conduct the Wilcoxon signed-rank test to compare the performance of our method with the other few-shot baselines, as shown in Tab.~\ref{tab:wilcoxon}. The results show statistically significant differences with $p < 0.05$, corresponding to a 95\% confidence level. 
\begin{table*}[h!]
\centering  
  \caption{Performance comparison across 10 datasets in few-shot scenarios.  ``*'' indicates performance without using feature weights. ``N/A'': not available due to token length. ``Ex.'':  the example-based prompt, ``E2E'': end-to-end inference with LLM during testing, and ``FT'': fine-tuning LLM. Since most baselines do not support the Myocardial dataset, the average AUC is computed over the remaining datasets.}
  \label{tab: sub_main_results}
\resizebox{\linewidth}{!}{
\begin{tabular}{c|c|ccc|cccccccccccc}
\cmidrule{1-17}\multirow{2}[3]{*}{Shot} & \multirow{2}[3]{*}{Method} & \multicolumn{3}{c|}{Requirement} & \multicolumn{11}{c}{Datasets}                                                         &  \\
\cmidrule{3-17}      &       & Ex.   & E2E   & FT    & Adult & Bank  & Blood & Car   & Credit-g & Cultivars & Diabetes & Heart & Myocardial & NHANES & Average (w/o Myocardial) & Average Rank \\
\midrule
\multirow{7}[0]{*}{4} & In-context & \checkmark   & \checkmark   & -    & 77.51 & 61.38 & 56.3  & 62.47 & 52.99 & 51.38 & 71.71 & 60.76 & N/A   & 91.84 & 65.15 & 5.67 \\
      & TABLET & \checkmark   & \checkmark   & -    & 75.29 & 58.11 & 56.45 & 60.21 & 54.33 & 54.28 & 63.96 & 68.19 & N/A   & 93.54 & 64.93 & 5.44 \\
      & LIFT  & \checkmark   & \checkmark   & -    & 83.07 & \underline{71.9} & \underline{69.86} & 70.16 & 52.89 & \underline{57.61} & 77.69 & 73.73 & N/A   & 90.19 & 71.9  & 3.67 \\
      & P2T   & \checkmark   & \checkmark   & -    & 82.66 & 70.5  & 63.69 & 50.36 & 49.6  & 51.8  & 73.01 & \underline{74.97} & N/A   & 97.96 & 68.28 & 4.56 \\
      & TabLLM & -    & \checkmark   & \checkmark   & 83.57 & 62.51 & 55.87 & \textbf{85.82} & 51.9  & 54.39 & 70.42 & 59.74 & N/A   & \textbf{99.49} & 69.3  & 4.44 \\
      & FeatLLM & \checkmark   & -    & -    & \textbf{86.68} & 70.45 & 68.34 & 72.69 & \underline{55.94} & 55.63 & \underline{80.28} & \textbf{75.66} & \underline{52.87} & 92.2  & 73.1  & 2.6 \\
      & \cellcolor[rgb]{ .906,  .902,  .902}Ours & \cellcolor[rgb]{ .906,  .902,  .902}- & \cellcolor[rgb]{ .906,  .902,  .902}- & \cellcolor[rgb]{ .906,  .902,  .902}- & \cellcolor[rgb]{ .906,  .902,  .902}\underline{83.74} & \cellcolor[rgb]{ .906,  .902,  .902}\textbf{82.53} & \cellcolor[rgb]{ .906,  .902,  .902}\textbf{76.31} & \cellcolor[rgb]{ .906,  .902,  .902}\underline{78.88} & \cellcolor[rgb]{ .906,  .902,  .902}\textbf{61.43} & \cellcolor[rgb]{ .906,  .902,  .902}\textbf{57.73} & \cellcolor[rgb]{ .906,  .902,  .902}\textbf{80.45} & \cellcolor[rgb]{ .906,  .902,  .902}74.12 & \cellcolor[rgb]{ .906,  .902,  .902}\textbf{63.25*} & \cellcolor[rgb]{ .906,  .902,  .902}\underline{98.1} & \cellcolor[rgb]{ .906,  .902,  .902}\textbf{77.03} & \cellcolor[rgb]{ .906,  .902,  .902}\textbf{1.5} \\
\midrule
\multirow{7}[0]{*}{8} & In-context & \checkmark   & \checkmark   & -    & 79.3  & 69.57 & 58.99 & 67.57 & 52.43 & 51.68 & 72.21 & 65.46 & N/A   & 86.67 & 67.1  & 5.67 \\
      & TABLET & \checkmark   & \checkmark   & -    & 77.56 & 69.08 & 56.37 & 65.53 & 52.9  & 51.48 & 65.47 & 69.85 & N/A   & 94.25 & 66.94 & 5.78 \\
      & LIFT  & \checkmark   & \checkmark   & -    & 83.58 & 66.56 & 64.4  & 76.72 & 53.34 & \underline{58.65} & 77.02 & 76.3  & N/A   & 89.89 & 71.83 & 3.89 \\
      & P2T   & \checkmark   & \checkmark   & -    & 83.56 & 66.75 & 62.52 & 54.46 & 50.92 & 52.58 & 74.95 & 77.01 & N/A   & 98.15 & 68.99 & 4.78 \\
      & TabLLM & -    & \checkmark   & \checkmark   & 83.52 & 63.19 & 66.01 & \textbf{87.43} & 56.42 & 52.86 & 64.3  & 70.14 & N/A   & \textbf{100} & 71.54 & 4 \\
      & FeatLLM & \checkmark   & -    & -    & \textbf{87.89} & \underline{75.85} & \underline{70.37} & 73.26 & \underline{57.42} & 56.97 & \underline{79.38} & \textbf{79.46} & \underline{56.22} & 93.29 & 74.88 & 2.4 \\
      & \cellcolor[rgb]{ .906,  .902,  .902}Ours & \cellcolor[rgb]{ .906,  .902,  .902}- & \cellcolor[rgb]{ .906,  .902,  .902}- & \cellcolor[rgb]{ .906,  .902,  .902}- & \cellcolor[rgb]{ .906,  .902,  .902}\underline{83.95} & \cellcolor[rgb]{ .906,  .902,  .902}\textbf{82.71} & \cellcolor[rgb]{ .906,  .902,  .902}\textbf{76.39} & \cellcolor[rgb]{ .906,  .902,  .902}\underline{79.67} & \cellcolor[rgb]{ .906,  .902,  .902}\textbf{63.26} & \cellcolor[rgb]{ .906,  .902,  .902}\textbf{59.35} & \cellcolor[rgb]{ .906,  .902,  .902}\textbf{80.48} & \cellcolor[rgb]{ .906,  .902,  .902}\underline{77.52} & \cellcolor[rgb]{ .906,  .902,  .902}\textbf{63.62*} & \cellcolor[rgb]{ .906,  .902,  .902}\underline{98.34} & \cellcolor[rgb]{ .906,  .902,  .902}\textbf{77.96} & \cellcolor[rgb]{ .906,  .902,  .902}\textbf{1.4} \\
\midrule
\multirow{7}[0]{*}{16} & In-context & \checkmark   & \checkmark   & -    & 79.5  & 69.76 & 56.59 & 76.94 & 55.29 & 54.31 & 71.64 & 67    & N/A   & 93.33 & 69.37 & 5.78 \\
      & TABLET & \checkmark   & \checkmark   & -    & 79.74 & 69.4  & 60.62 & 74.02 & 51.65 & 57.44 & 66.71 & 68.39 & N/A   & 95.02 & 69.22 & 5.56 \\
      & LIFT  & \checkmark   & \checkmark   & -    & 84.96 & 69.97 & 56.71 & 77.98 & 51.03 & \textbf{62.92} & 74.3  & 80.43 & N/A   & 89.21 & 71.95 & 4.33 \\
      & P2T   & \checkmark   & \checkmark   & -    & \underline{86.02} & 72.02 & 62.12 & 61    & 54.05 & 53.52 & 73.35 & 80.06 & N/A   & 97.71 & 71.09 & 4.44 \\
      & TabLLM & -    & \checkmark   & \checkmark   & 83.23 & 63.73 & 65.14 & \textbf{88.65} & \underline{60.38} & 56.97 & 67.34 & \underline{81.72} & N/A   & \textbf{100} & 74.13 & 3.56 \\
      & FeatLLM & \checkmark   & -    & -    & \textbf{87.54} & \underline{78.41} & \underline{70.07} & 79.43 & 56.6  & 57.19 & \underline{80.15} & \textbf{83.71} & \underline{55.32} & 95.64 & 76.53 & 2.4 \\
      & \cellcolor[rgb]{ .906,  .902,  .902}Ours & \cellcolor[rgb]{ .906,  .902,  .902}- & \cellcolor[rgb]{ .906,  .902,  .902}- & \cellcolor[rgb]{ .906,  .902,  .902}- & \cellcolor[rgb]{ .906,  .902,  .902}84 & \cellcolor[rgb]{ .906,  .902,  .902}\textbf{83.13} & \cellcolor[rgb]{ .906,  .902,  .902}\textbf{75.98} & \cellcolor[rgb]{ .906,  .902,  .902}\underline{81.02} & \cellcolor[rgb]{ .906,  .902,  .902}\textbf{65.22} & \cellcolor[rgb]{ .906,  .902,  .902}\underline{60.45} & \cellcolor[rgb]{ .906,  .902,  .902}\textbf{80.46} & \cellcolor[rgb]{ .906,  .902,  .902}81.56 & \cellcolor[rgb]{ .906,  .902,  .902}\textbf{64.03*} & \cellcolor[rgb]{ .906,  .902,  .902}\underline{99.09} & \cellcolor[rgb]{ .906,  .902,  .902}\textbf{78.99} & \cellcolor[rgb]{ .906,  .902,  .902}\textbf{1.8} \\
\midrule
\multirow{7}[0]{*}{32} & In-context & \checkmark   & \checkmark   & -    & 81.89 & 66.93 & 58.69 & 81.64 & N/A   & N/A   & 73.32 & 71.94 & N/A   & 88.54 & N/A   & 5.43 \\
      & TABLET & \checkmark   & \checkmark   & -    & 78.08 & 73.61 & 57.94 & 76.44 & N/A   & N/A   & 66.97 & 71.9  & N/A   & 95.82 & N/A   & 5.86 \\
      & LIFT  & \checkmark   & \checkmark   & -    & \underline{84.6} & 69.29 & 56.51 & 78.34 & N/A   & 54.7  & 74.14 & 81.27 & N/A   & 86.93 & N/A   & 5 \\
      & P2T   & \checkmark   & \checkmark   & -    & 84.11 & 72.21 & 67.68 & 62.48 & N/A   & \underline{61.39} & 75.94 & 82.85 & N/A   & 98.73 & N/A   & 3.75 \\
      & TabLLM & -    & \checkmark   & \checkmark   & 82.6  & 66.51 & 69.95 & \textbf{89.02} & \underline{68.64} & 58.5  & 69.74 & \textbf{87.43} & N/A   & \textbf{100} & 76.93 & 3.33 \\
      & FeatLLM & \checkmark   & -    & -    & \textbf{87.09} & \underline{78.37} & \underline{71.13} & \underline{85.01} & 61.79 & 59.62 & \underline{80.06} & \underline{87.19} & \underline{60.02} & 97.29 & \underline{78.62} & \underline{2.3} \\
      & \cellcolor[rgb]{ .906,  .902,  .902}Ours & \cellcolor[rgb]{ .906,  .902,  .902}- & \cellcolor[rgb]{ .906,  .902,  .902}- & \cellcolor[rgb]{ .906,  .902,  .902}- & \cellcolor[rgb]{ .906,  .902,  .902}83.97 & \cellcolor[rgb]{ .906,  .902,  .902}\textbf{83.86} & \cellcolor[rgb]{ .906,  .902,  .902}\textbf{76.16} & \cellcolor[rgb]{ .906,  .902,  .902}82.42 & \cellcolor[rgb]{ .906,  .902,  .902}\textbf{69.03} & \cellcolor[rgb]{ .906,  .902,  .902}\textbf{62.01} & \cellcolor[rgb]{ .906,  .902,  .902}\textbf{81.02} & \cellcolor[rgb]{ .906,  .902,  .902}83.94 & \cellcolor[rgb]{ .906,  .902,  .902}\textbf{65.44*} & \cellcolor[rgb]{ .906,  .902,  .902}\underline{99.42} & \cellcolor[rgb]{ .906,  .902,  .902}\textbf{80.20} & \cellcolor[rgb]{ .906,  .902,  .902}\textbf{1.8} \\

\midrule
\multirow{7}[0]{*}{64} & In-context & \checkmark   & \checkmark   & -    & N/A   & N/A   & 65.79 & 77.65 & N/A   & N/A   & 70.22 & N/A   & N/A   & N/A   & N/A   & 4.67 \\
      & TABLET & \checkmark   & \checkmark   & -    & N/A   & N/A   & 63.47 & 76.13 & N/A   & N/A   & 69.27 & N/A   & N/A   & N/A   & N/A   & 5.67 \\
      & LIFT  & \checkmark   & \checkmark   & -    & N/A   & N/A   & 58.79 & 83.1  & N/A   & N/A   & N/A   & N/A   & N/A   & N/A   & N/A   & 5.5 \\
      & P2T   & \checkmark   & \checkmark   & -    & N/A   & N/A   & 67.94 & N/A   & N/A   & N/A   & N/A   & N/A   & N/A   & N/A   & N/A   & 4 \\
      & TabLLM & -    & \checkmark   & \checkmark   & \underline{84.88} & 70.83 & 70.88 & \textbf{92.18} & \underline{70.8}  & \underline{60.32} & \underline{71.56} & \textbf{89.78} & N/A   & \textbf{100} & 79.03 & 2 \\
      & FeatLLM & \checkmark   & -    & -    & \textbf{87.77} & \underline{81.18} & \underline{71.04} & \underline{86.78} & 66.43 & 59.14 & 80.91 & \underline{88.08} & \underline{61.47} & 98.32 & \underline{79.96} & 2.2 \\
      & \cellcolor[rgb]{ .906,  .902,  .902}Ours & \cellcolor[rgb]{ .906,  .902,  .902}- & \cellcolor[rgb]{ .906,  .902,  .902}- & \cellcolor[rgb]{ .906,  .902,  .902}- & \cellcolor[rgb]{ .906,  .902,  .902}84.25 & \cellcolor[rgb]{ .906,  .902,  .902}\textbf{84.52} & \cellcolor[rgb]{ .906,  .902,  .902}\textbf{75.97} & \cellcolor[rgb]{ .906,  .902,  .902}84.1  & \cellcolor[rgb]{ .906,  .902,  .902}\textbf{72.6} & \cellcolor[rgb]{ .906,  .902,  .902}\textbf{65.09} & \cellcolor[rgb]{ .906,  .902,  .902}\textbf{81.5} & \cellcolor[rgb]{ .906,  .902,  .902}86.25 & \cellcolor[rgb]{ .906,  .902,  .902}\textbf{65.75*} & \cellcolor[rgb]{ .906,  .902,  .902}99.65 & \cellcolor[rgb]{ .906,  .902,  .902}\textbf{81.55} & \cellcolor[rgb]{ .906,  .902,  .902}\textbf{1.7} \\
\bottomrule
\end{tabular}%
}
\end{table*}

\begin{table*}[t]
    \centering
    \caption{Wilcoxon signed-rank test between ProtLLM and other baselines. }
     \resizebox{0.8\linewidth}{!}{
\begin{tabular}{c|ccccccrcc}
\toprule
Model & LogReg & TabPFN & STUNT & In-context & TABLET & LIFT  & \multicolumn{1}{l}{P2T} & TabLLM & FeatLLM \\
\midrule
p     & 4.44E-09 & 7.95E-06 & 6.71E-08 & 1.46E-11 & 1.46E-11 & 7.80E-09 & 1.86E-08 & 5.42E-04 & 1.05E-05 \\
\bottomrule
\end{tabular}%
    }
    \label{tab:wilcoxon}
\end{table*}



Furthermore, we include a comparison against non-LLM methods in Fig.~\ref{fig:non-llm-results} to comprehensively assess its performance. Clearly, ours is superior to non-LLM methods, including TabPFN\_v2 and STUNT, the popular few-shot methods. Both STUNT and ours consider prototype learning, where we provides a different training-free way by using LLM for prototype estimation. It thus proves the effectiveness of ours. With the development of shot, i.e., the number of labeled training samples, the performance gap between ours and baselines is becoming gradually small, no matter for LLM-based or non-LLM-based baselines. It is reasonable since we design the example-free prompt for LLM and a training-free prototype estimation framework, making ours more suitable for zero-shot and very few-shot settings. When training samples become more and more, prompting LLM with selected high-quality examples or training a classifier might be a suitable way for taking full advantage of samples. Detailed results are presented in Tab.~\ref{tab:non-LLM-based methods results}
of Appendix~\ref{appendix:detail_results}.

\subsection{Further Discussion}
\textbf{{Ablation Study. }} We aim to identify which components of ProtoLLM play a central role in prompting LLM in tabular data learning, including example-free prompt, decomposition into sub-problems, and feature weighting. We consider three variants: (1) replace our example-free prompt with an example-based prompt by including few-shot tabular examples;
(2) generate all $D$ features simultaneously instead of decomposing them into sub-problems;
(3) remove feature weighting and treat all features as equally important.
Tab.~\ref{tab:ablation} summarizes the results across all datasets except Myocardial, while Tab.~\ref{tab: results of different generation types}
in the Appendix~\ref{appendix:detail_results}
presents more detailed results for selected datasets. 

Across datasets, within our experimental setting, the example-free prompt consistently outperforms the example-based one. This indicates that, under the low-shot scenarios we consider, avoiding the use of examples enables more effective utilization of the large model’s prior knowledge. Regarding generation types, without examples, decomposing the task into sub-problems by generating features individually usually outperforms generating all features simultaneously, as individual generation better captures fine-grained feature–target relationships that joint generation may dilute.
When examples are provided, the advantage of decomposing the task lessens, since examples already convey some inter-feature relationships, which benefits joint feature generation. Adding feature weighting further improves classification in the example-free setting with decomposed feature generation, indicating that assigning different importance to features helps build better prototypes. Weighting also provides some benefit when examples are used with decomposed feature generation, but its effect is minimal in the example-based setting with joint feature generation, since the joint generation and example information already sufficiently capture inter-feature dependencies, leaving limited room for weighting to contribute.

Overall, our full ProtoLLM method, which combines example-free prompting, decomposition into sub-problems, and feature weighting, achieves the best results. This confirms its effectiveness in leveraging prior knowledge and modeling feature importance for low-shot tabular learning.

\begin{table}[htbp]
    \centering
    \caption{Average AUC scores with different variants across datasets without myocardial. \texttt{F} denotes feature-level generation and \texttt{D} denotes decomposition into sub-problems. \texttt{E} denotes example-based prompt for LLMs.}
    \label{tab:ablation}
    \resizebox{0.9\linewidth}{!}{
    \begin{tabular}{c|c|c|cccc}
    \toprule
    \multirow{2}[4]{*}{E} & \multirow{2}[4]{*}{D} & \multirow{2}[4]{*}{W} & \multicolumn{4}{c}{shot} \\
    \cmidrule{4-7}      
    &       &       & 0     & 4     & 8     & 16 \\
    \midrule
    \rowcolor{red!20}
    {-} & {\checkmark} & \multicolumn{1}{l|}{\checkmark} & \textbf{76.09 } & \textbf{77.03 } & \textbf{77.96 } & \textbf{78.99 } \\
    \rowcolor{red!20}
    {-}      &  {\checkmark}     &   -    & \underline{73.59 } & \underline{75.17 } & \underline{76.36 } & \underline{78.04 } \\
    \rowcolor{yellow!20}
    {-} & {-} & \multicolumn{1}{l|}{\checkmark} & 73.13  & 72.03  & 74.83  & 77.13  \\
    \rowcolor{yellow!20}
    {-}      &  {-}     &   -    & 73.22  & 72.27  & 74.80  & 77.49  \\
    \rowcolor{green!20}
    {\checkmark} & {\checkmark} & \multicolumn{1}{l|}{\checkmark} &  N/A     & 71.63  & 73.46  & 75.83  \\
    \rowcolor{green!20}
    {\checkmark}       &   {\checkmark}    &   -    &   N/A  & 67.27  & 70.99  & 74.54  \\
    \rowcolor{blue!20}
    {\checkmark} & {-} & \multicolumn{1}{l|}{\checkmark} &   N/A    & 70.66  & 73.61  & 75.92  \\
    \rowcolor{blue!20}
    {\checkmark}       &  {-}     &    -   &  N/A     & 70.96  & 73.79  & 76.58  \\
    \bottomrule
    \end{tabular}%
    }
\end{table}

\begin{figure}[h]
\centering

\includegraphics[width=\columnwidth]{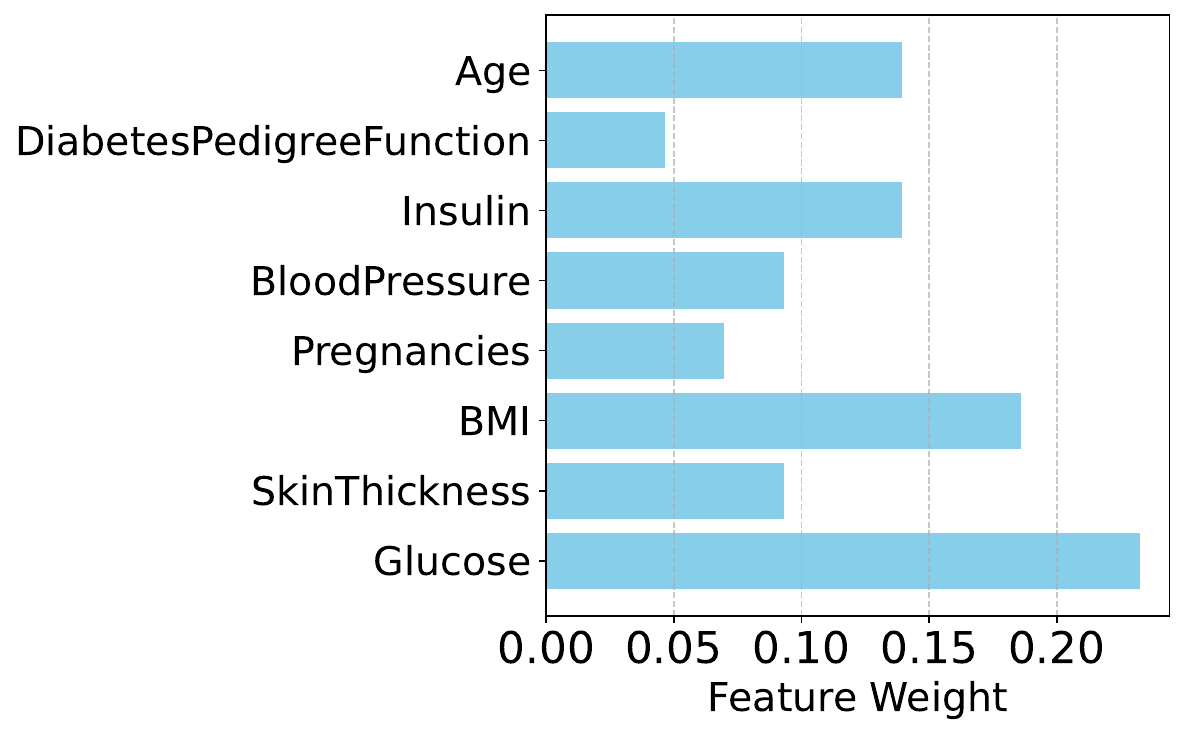}

\caption{Visualization of feature weights derived from the LLM on the \textit{Diabetes} dataset. The most important features identified include \textit{Glucose}, \textit{BMI}, \textit{Age}, and \textit{Insulin}, which are well-aligned with known clinical risk factors for diabetes.}

\label{fig:diabetes_weight}
\end{figure}

\textbf{Feature Weights.} We visualize the feature weights derived from LLM on diabetes dataset in  Fig.~\ref{fig:diabetes_weight}. This dataset aims to predict the likelihood of an individual having diabetes based on various health-related features. As expected, \textit{Glucose} is the most influential feature associated with diabetes, as blood glucose levels directly reflect the body's ability to regulate sugar. Other important features, including  \textit{BMI}, and \textit{Age}, and \textit{Insulin}, also play significant roles due to their established associations with diabetes.Notably, these LLM-derived weights align with established medical knowledge, including key risk factors identified by the WHO, such as elevated blood glucose, overweight, and abnormal insulin levels. This agreement supports both the interpretability and practical value of our approach.

\textbf{Impacts of Base LLMs.} Note that \textsc{ProtoLLM} is compatible with various LLMs for enhancing tabular data analysis. We evaluate both closed-source models (GPT-3.5 and GPT-4o) and open-source models (LLaMA-3B and LLaMA-8B), as shown in Tab.~\ref{tab:Other_LLMs}. While performance generally improves with stronger models, \textsc{ProtoLLM} remains functional and effective even when used with smaller models like LLaMA-3B. This indicates that our approach does not rely on the implicit capacity of large models alone, but provides a robust framework that can benefit from stronger reasoning and generation abilities when available.


\begin{table}[h]
    \centering
    \caption{Comparison of AUC scores with different LLMs.}
    \resizebox{0.95\linewidth}{!}{
\begin{tabular}{c|c|cccc}
\toprule
dataset & shot  & Llama3B & Llama8B & GPT-3.5 & GPT-4o \\
\midrule
\multirow{4}[2]{*}{Adult} & 0     & 78.20 & 80.90 & \underline{83.7} & \textbf{84.30} \\
      & 4     & 78.37 & 81.03 & \underline{83.74} & \textbf{84.49} \\
      & 8     & 78.64 & 81.00 & \underline{83.95} & \textbf{84.55} \\
      & 16    & 79.42 & 81.18 & \underline{84.00} & \textbf{84.70} \\
\midrule
\multirow{4}[2]{*}{Bank} & 0     & 54.48 & 62.00 & \underline{82.93} & \textbf{84.29} \\
      & 4     & 56.06 & 63.43 & \underline{82.53} & \textbf{84.02} \\
      & 8     & 56.80 & 64.75 & \underline{82.71} & \textbf{84.19} \\
      & 16    & 58.52 & 66.86 & \underline{83.13} & \textbf{84.28} \\
\midrule
\multirow{4}[2]{*}{Blood} & 0     & 74.90 & 74.64 & \underline{75.31} & \textbf{76.43} \\
      & 4     & 75.61 & 74.56 & \textbf{76.31} & \underline{76.11} \\
      & 8     & 75.57 & 74.57 & \textbf{76.39} & \underline{75.82} \\
      & 16    & \textbf{76.43} & 73.99 & \underline{75.98} & 75.71 \\
\midrule
\multirow{4}[2]{*}{Car} & 0     & 69.70 & 70.73 & \underline{77.76} & \textbf{80.59} \\
      & 4     & 71.29 & 71.11 & \underline{78.88} & \textbf{81.46} \\
      & 8     & 72.05 & 71.50 & \underline{79.67} & \textbf{82.19} \\
      & 16    & 74.23 & 73.07 & \underline{81.02} & \textbf{83.38} \\
\bottomrule
\end{tabular}%
}
    \label{tab:Other_LLMs}
\end{table}

\begin{figure}[h]
\centering

\includegraphics[width=\columnwidth]{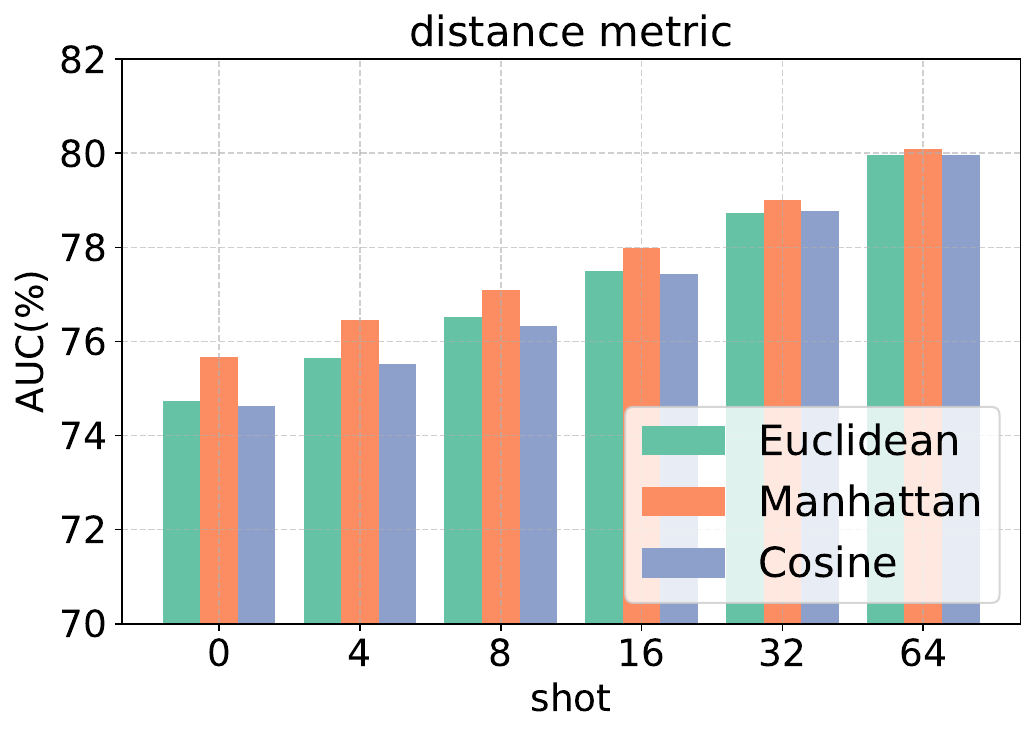}

\caption{Average performance across all datasets using different distance metrics: Euclidean, Manhattan, and Cosine.}

\label{fig:distance metrics}
\end{figure}

\textbf{Impacts of Distance Metrics.}
We investigate the impact of different distance metrics on ProtoLLM's performance. As shown in Fig.~\ref{fig:distance metrics}, we report the average AUC across ten datasets under varying shot settings. As the number of shots increases, the performance gap between different distance metrics tends to narrow. Overall, all commonly used metrics achieve strong results, with Manhattan distance  performing even better than the default Euclidean. Detailed results on individual datasets are provided in Tab.~\ref{tab: distance metrics}
of Appendix~\ref{appendix:detail_results}.
The optimal distance metric varies across datasets, reflecting the heterogeneity of tabular data. Different similarity measures may be more suitable depending on the task, and ProtoLLM's design allows flexible replacement of the distance function to further improve performance.

\textbf{Weight Normalization.}
We compare different normalization methods for the LLM-generated feature weights to assess their effect on prototype robustness and model performance. Specifically, we evaluate: (1) softmax normalization with different temperature values, where lower temperatures (e.g., $T=0.1$) emphasize large weights and higher temperatures (e.g., $T=2.0$) yield more uniform distributions; (2) min-max normalization, which linearly rescales weights to $[0,1]$; and (3) sum normalization (ours), which normalizes by the total weight sum. Tab.~\ref{tab:weight_norm} shows the average AUC across datasets. Softmax with $T=0.5$ achieves the best performance, balancing feature emphasis and stability. Our simple sum normalization also performs well without requiring hyperparameter selection.

\begin{table}[h]
    \centering
    \caption{Results of different normalization method of weights.}
\begin{tabular}{c|cccc}
\toprule
normalization-method & shot-0 & shot-4 & shot-8 & shot-16 \\
\midrule
softmax-0.1 & 73.63 & 74    & 74.51 & 75 \\
softmax-0.5 & \textbf{76.24} & \textbf{77.31} & \textbf{78.23} & \underline{79.25} \\
softmax-1.0 & 75.89 & \underline{77.22} & \textbf{78.23} & \textbf{79.47} \\
softmax-2.0 & 75.09 & 76.58 & 77.67 & 79.08 \\
Min-Max & 75.45 & 76.35 & 77.24 & 78.26 \\
\rowcolor{gray!20}
sum-normalization & \underline{76.09} & 77.03 & 77.96 & 78.99 \\
\bottomrule
\end{tabular}%
    
    \label{tab:weight_norm}
\end{table}

\textbf{Impact on Prompt Design.}
To examine the robustness of ProtoLLM to prompt wording, we performed an ablation study by modifying different components of the original prompt. Three prompt variants were considered:

\begin{itemize}
    \item \textbf{Altered Task Instruction}: The original task description \textit{``Given a feature description and a task, your goal is...''} was replaced with \textit{``You are asked to examine the connection between a provided feature and a task's target variable, and then generate a dictionary containing the relevant details.''}.
    \item \textbf{Simplified CoT Reasoning Instruction}: The original chain-of-thought reasoning guide \textit{``First, thoroughly analyze... Then, based on this analysis, create a dictionary...''} was replaced with a minimal prompt \textit{``Let's think step by step.''}.
    \item \textbf{Altered Output Format}: The output format was changed from a JSON snippet to a Python code block that produces the same dictionary object.
\end{itemize}

Tab.~\ref{tab:prompt_ablation} presents the results on three representative datasets. The original prompt yields performance that is consistently on par with, and sometimes better than, its variants, indicating its suitability as a default configuration.
\begin{table}[htbp]
\centering
\caption{Performance comparison of different prompt variants.}
\label{tab:prompt_ablation}
\begin{tabular}{c|c|cccc}
\toprule
\multirow{4}[2]{*}{adult} & Altered Task Instruction & \underline{83.71} & \underline{83.72} & \textbf{83.95} & \underline{83.87} \\
& Simplified CoT Reasoning & 83.08 & 83.19 & 83.46 & 83.36 \\
& Altered Output Format & 83.33 & 83.36 & 83.66 & 83.51 \\
& \textbf{Original Prompt} & \textbf{83.72} & \textbf{83.74} & \textbf{83.95} & \textbf{84.00} \\
\midrule
\multirow{4}[2]{*}{bank} & Altered Task Instruction & \underline{80.57} & 79.90 & 79.84 & 80.49 \\
& Simplified CoT Reasoning & 81.07 & \underline{80.19} & \underline{79.92} & \underline{80.58} \\
& Altered Output Format & 79.23 & 78.55 & 78.65 & 79.68 \\
& \textbf{Original Prompt} & \textbf{82.93} & \textbf{82.53} & \textbf{82.71} & \textbf{83.13} \\
\midrule
\multirow{4}[2]{*}{blood} & Altered Task Instruction & \underline{75.79} & 75.23 & 74.80 & 74.77 \\
& Simplified CoT Reasoning & 74.85 & 74.50 & 74.44 & 74.48 \\
& Altered Output Format & \textbf{76.77} & \underline{76.04} & \underline{75.51} & \underline{75.42} \\
& \textbf{Original Prompt} & 75.31 & \textbf{76.31} & \textbf{76.39} & \textbf{75.98} \\
\bottomrule
\end{tabular}%
\end{table}

\begin{figure}[h]
\centering

\includegraphics[width=0.9\linewidth]{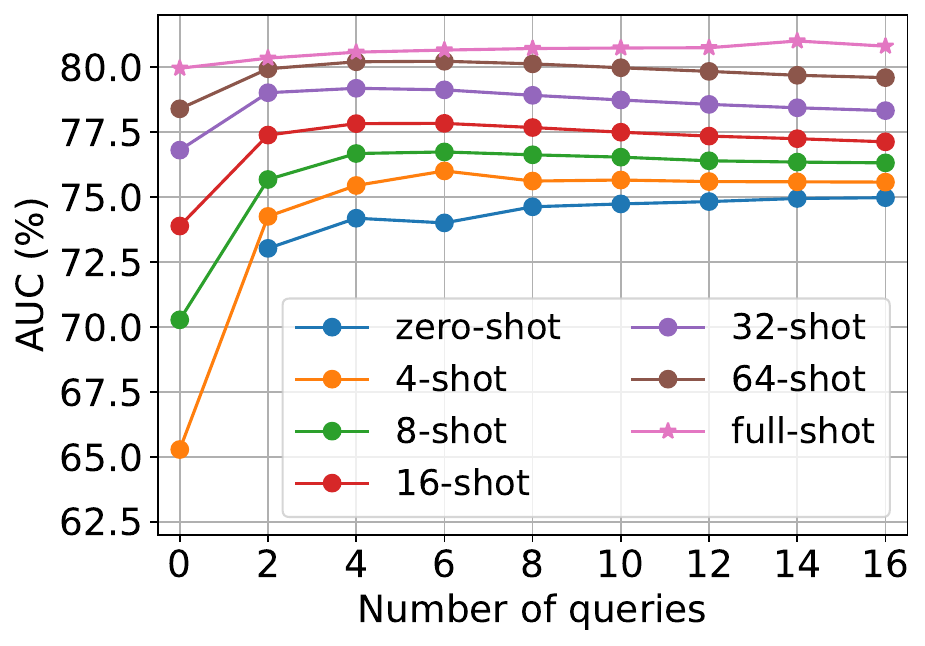}
\caption{Average AUC of \textsc{ProtoLLM} under different numbers of LLM queries $K$ across ten datasets.}

\label{fig:AUC-queries}
\end{figure}

\textbf{Analysis of Query Times.}
We query LLMs $K$ times for better feature value generation in previous experiments. We here test the impact of $K$ and report the average results on 10 datasets in Fig.~\ref{fig:AUC-queries}. Generally, $K$ balances the weights of prior knowledge and data likelihood. $K = 0$ in the X-axis means that we only use the training samples to construct the prototype. We can find that, for different few-shot settings, the performance of $K = 0$ is lower than that of $K > 0$, showing the effectiveness of the prototype built by the LLM. As $K$ increases, LLMs become more significant in the final prototype construction. One can obtain higher results after selecting the optimal $K$ on the validation datasets. We report the performance of full-shot with $K=0$. ProtoLLM with few-shot samples achieves the desired performance when compared to the full-shot, where 64-shot is very close to full-shot.

\textbf{Token Consumption.}
In Tab.~\ref{tab:token_counts}, we compare the average token consumption per query of our method with that of FeatLLM, which leverages large language models (LLMs) to extract both rules and functions. Benefiting from its example-free prompt design and feature-level querying strategy, ProtoLLM consistently requires significantly fewer tokens than FeatLLM. This improved token efficiency not only reduces computational overhead but also facilitates faster inference and enhances scalability.

\begin{table}[h]
    \centering
    \caption{Average number of tokens per query.}
    \resizebox{\linewidth}{!}{
    \begin{tabular}{c|c|cccc}
    \toprule
          &  Type     & \multicolumn{1}{l}{Adult} & \multicolumn{1}{l}{Bank} & \multicolumn{1}{l}{Blood} & \multicolumn{1}{l}{Car} \\
    \midrule
    \multirow{2}[2]{*}{FeatLLM} & FeatLLM-rule & 1454.2  & 1324.8  & 581.0  & 693.4  \\
          & FeatLLM-function & 710.8  & 531.8  & 304.4  & 376.8  \\
    \midrule
    \multirow{2}[2]{*}{ProtoLLM} & ProtoLLM-weights & 445.0  & 423.0  & 216.0  & 259.0  \\
          & ProtoLLM-feature & 245.2  & 223.1  & 208.8  & 278.5  \\
    \bottomrule
    \end{tabular}%

}
    \label{tab:token_counts}
\end{table}

\textbf{Time Consumption.} 
By restricting LLM access to the \textit{Pre-Test Phase} only, both ProtoLLM and FeatLLM reduce inference overhead. ProtoLLM, in particular, achieves lower computation time on datasets with fewer features, such as \textit{Blood} and \textit{Car}.
In contrast, methods like P2T, LIFT, and others that perform per-sample prompting must query the LLM individually for each test example, leading to significantly higher latency and token usage during inference. By avoiding any LLM interaction at test time, ProtoLLM achieves substantially faster inference and lower resource consumption, making it highly suitable for deployment in real-world, large-scale, or resource-constrained settings.

\begin{table}[h!]
  \centering
  \caption{Average time to access the LLM; \textit{0} indicates no LLM access required.}
    \begin{tabular}{c|c|cccc}
    \toprule
          & Model & Adult & Bank  & Blood & Car \\
    \midrule
    \multirow{4}[2]{*}{Pre-Test Phase} & LIFT  & 0     & 0     & 0     & 0 \\
          & P2T   & 6.0  & 12.0  & 5.9  & 5.8  \\
          & FeatLLM & 562.1  & 564.1  & 1470.2  & 1020.6  \\
          & ours  & 1564.1  & 1934.2  & 530.9  & 738.8  \\
    \midrule
    \multirow{4}[2]{*}{Test Phase} & LIFT  & 20866.7  & 22391.0  & 309.6  & 770.8  \\
          & P2T   & 27810.6  & 21714.4  & 297.9  & 807.0  \\
          & FeatLLM &   0   & 0     & 0     & 0 \\
          & ours  & 0     & 0     & 0     & 0 \\
    \bottomrule
    \end{tabular}%
  \label{tab:time_cost}%
\end{table}%

\begin{figure}[htbp]  
  \centering
  \includegraphics[width=0.99\linewidth]{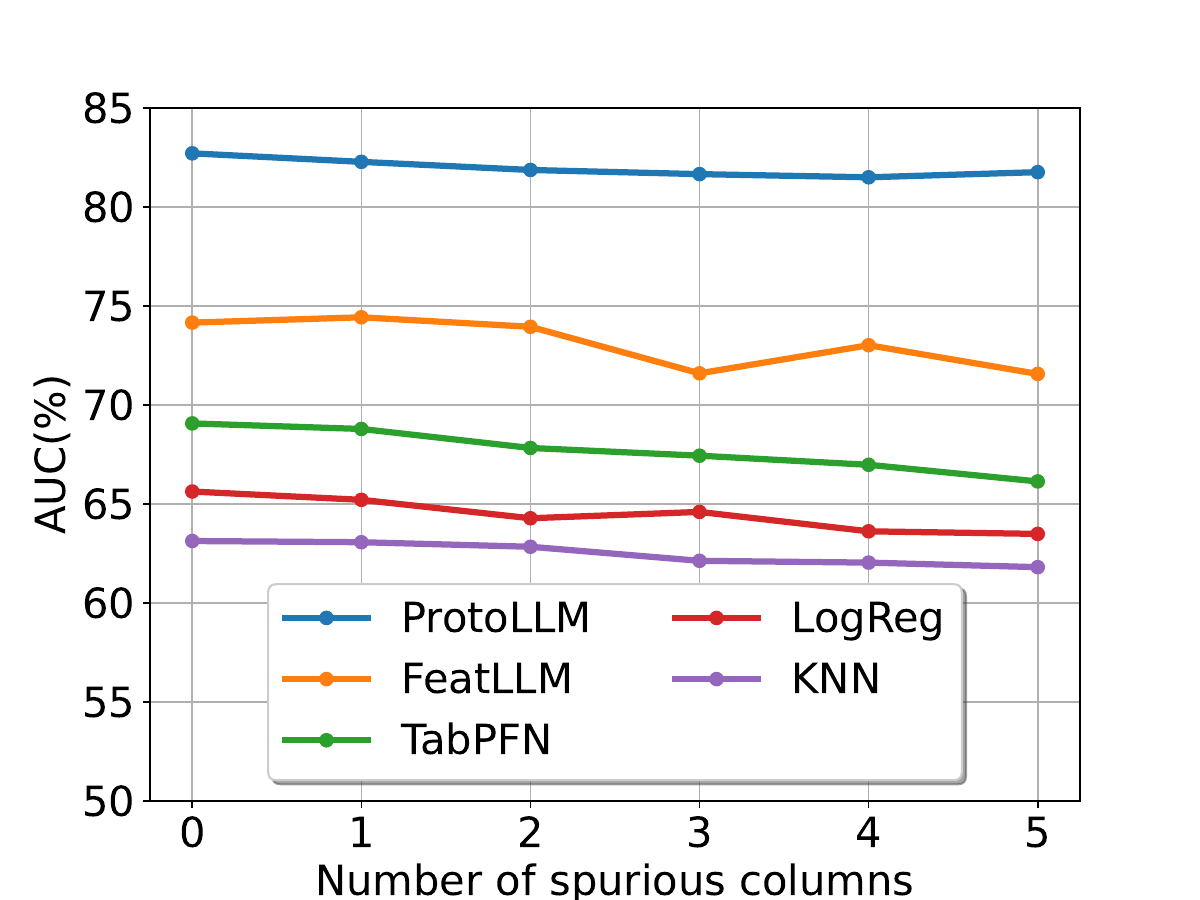}
  \caption{spurious-correlation analyse} 
  \label{fig:spurious-correlation}
\end{figure}

\textbf{Impact of Spurious Correlations.}
To analyze the impact of spurious correlations, following~\cite{han2024featllm}, we conduct experiments by randomly selecting columns from the Adult dataset and applying them to the Bank dataset. The Adult dataset predicts a person’s income and its features may introduce spurious correlations that could affect the task in Bank, which involves predicting the likelihood of a person subscribing to a term deposit. For this analysis, we set the number of shots to 4 and included 0-5 columns from the Adult dataset. The performance changes are shown in blue{ Fig.~\ref{fig:spurious-correlation}.} As the number of spurious columns increases, the performance of ProtoLLM remains stable. This is mainly because assigning weights to each feature, helps mitigate this effect and reduce the influence of spurious correlations.

\begin{figure}[htbp]
\centering

\includegraphics[width=0.9\columnwidth]{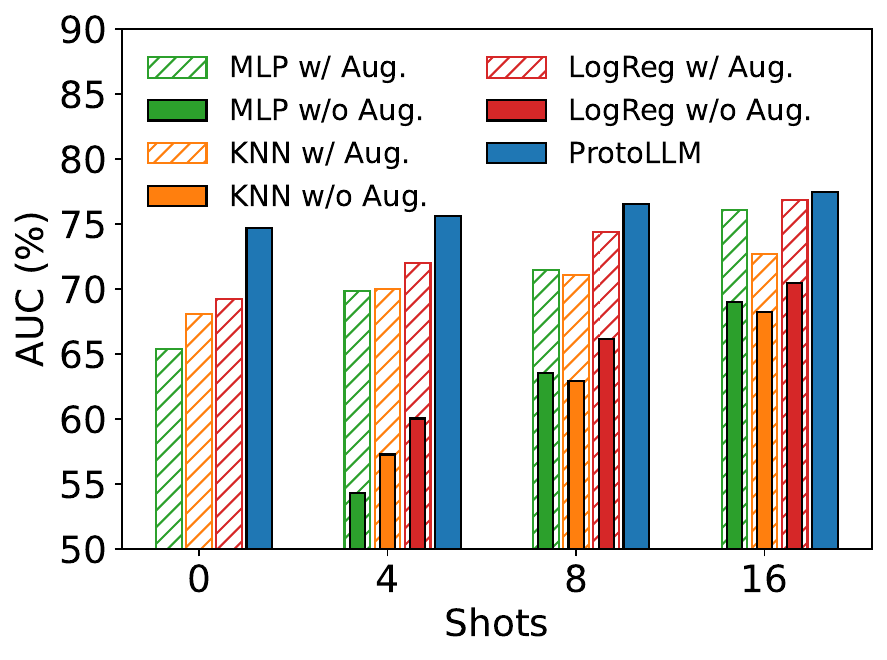}

\caption{Average AUC results of baselines with and without data augmentations by ProtoLLM.}
\label{fig:augmentation on tradition models}

\end{figure}

\textbf{ProtoLLM as Data Augmentation.} Previous results suggest that our training and example-free framework have great potential for high-quality generation.
Let $\hat{x}_{c}^{k} = \left[z_{c,1}^{k}, z_{c,2}^{k}, \ldots, z_{c,D}^{k} \right]$ denote $k$-th augmented sample for class $c$. To evaluate such data augmentation abilities of ProtoLLM, we first use these $K \times C$ samples to augment $\mathcal{S}$ and then apply traditional machine learning methods. The results in Fig.~\ref{fig:augmentation on tradition models} show that our augmented samples significantly improve the classification accuracy on LogReg, KNN, and MLP, verifying the superior quality of the data generated by our approach. However, despite these gains, they are still inferior to ours, demonstrating the advantage of prototype estimation in our approach. {Please see Tab.~\ref{tab:Data Augmentation}
in Appendix~\ref{appendix:detail_results}
for more details.}

\section{Conclusion}

We propose ProtoLLM, a training framework for zero and few-shot tabular data classification. This framework combines prior knowledge embedded in LLMs with data likelihood to construct class prototypes. We show that it is possible to efficiently draw out  common sense from LLMs and generate feature values by designing example-free prompts. We also showcase our ProtoLLM can be used as data augmentation and boosts traditional algorithms. Extensive experiments prove the superior performance of ours. Our ProtoLLM is LLM-agnostic and can be benefit from stronger LLM. The proposed method offers potential for future research in valuable insights into the utility of LLMs for tabular data analysis.




\bibliographystyle{IEEEtran}

\clearpage
\appendix
\section{Appendix / supplemental material}
\subsection{Example of Prompt and Answer}
\label{appendix: Example of prompt}
For numerical features, the prompts differ slightly. We require LLMs to infer their ranges. Additionally, we limit the length of the list due to the variations across different queries. Fig.~\ref{fig:prompt for regression} demonstrates a prompt used to generate feature values for the `age' attribute in the Adult dataset:\\
\begin{figure*}[b!]
    \centering
    
    \begin{tikzpicture}
        \node[draw, rounded corners=5pt, text width=\textwidth, align=left] (box) {
    You are an expert in analyzing relationships between feature and target variables. \\
    
    Given a feature description and a task, your goal is to analyze how the feature relates to the target and then generate a dictionary with specific details. \\

    \vspace{1em}
    Task: Does this person earn more than 50000 dollars per year? Yes or no? \\

    Feature: age, the age of an individual(age$>$16)(numeric variable, you should use your prior knowledge to determine the appropriate ranges of values) \\
    \vspace{1em}
    First conduct a thorough analysis of the relationship between the feature and task using your prior knowledge.  \\
    Then based on this analysis, create a dictionary with the following format: \\
    \vspace{1em}
    The output should be a markdown code snippet formatted in the following schema, including the leading and trailing \texttt{"}\texttt{```}json\texttt{"} and \texttt{"}\texttt{```}\texttt{"}:  \\
    \vspace{1em}
    \texttt{```}json  \\
    \{  \\
            \texttt{"}no\texttt{"}: [list of integers]  // possible values of feature age for target class no  \\
            \texttt{"}yes\texttt{"}: [list of integers]  // possible values of feature age for target class yes  \\
    \}  \\
    \texttt{```}  \\
    The age values should be presented as integers in lists.
    Make sure to include 5 typical age values for each target class in [\texttt{"}no\texttt{"}, \texttt{"}yes\texttt{"}].
    };
    \end{tikzpicture}
\caption{Prompt for querying  values of feature `age' in Adult dataset.}
\label{fig:prompt for regression}
\end{figure*}

\begin{figure*}[b!]
    \centering
    \begin{tikzpicture}
        \node[draw, rounded corners=5pt, text width=\textwidth, align=left] (box) {
    Based on prior knowledge, we can analyze the relationship between age and the task of determining if a person earns more than 50000 dollars per year. \\
    \vspace{1em}
    Typically, as an individual progresses in their career, their income tends to increase with age until they reach their peak earning potential. Therefore, it is reasonable to expect that people in older age groups are more likely to earn more than 50000 dollars per year compared to younger age groups. \\
    However, it is important to note that this is a general trend and there will always be exceptions. Other factors such as education level, occupation, and work experience can also im\vspace{1em}pact an individual's income. \\
    
    Based on this analysis, let's create a dictionary with the specified format: \\

    \texttt{```}json \\
    \{ \\
        \texttt{"}no\texttt{"}: [17, 18, 19, 20, 21],   \\
        \texttt{"}yes\texttt{"}: [35, 40, 45, 50, 55]  \\ 
    \} \\
    \texttt{```} \\

    This dictionary includes 5 typical age values for each target class, where "no" represents individuals earning less than or equal to 50000 dollars per year, and "yes" represents individuals earning more than 50000 dollars per year. The age values provided are just examples and can be adjusted based on the specific dataset and its distribution. \\
    };
    \end{tikzpicture}
    \caption{Answer for querying values of feature `age’ in Adult dataset.}
    \label{fig:answer for regression}
\end{figure*}

The answer for feature `age' is detailed in Fig.~\ref{fig:answer for regression}, where LLMs analyze the relationship and provide responses for various values of the `age' feature corresponding to each class in the Adult dataset.\\

\subsection{Other Tasks}
\label{appendix: Other tasks}
\subsubsection{Regression Task}
\label{appendix: Regression}

\begin{figure*}[!htbp]
\centering

    \subfloat[prompt]{\includegraphics[width=0.99\columnwidth]{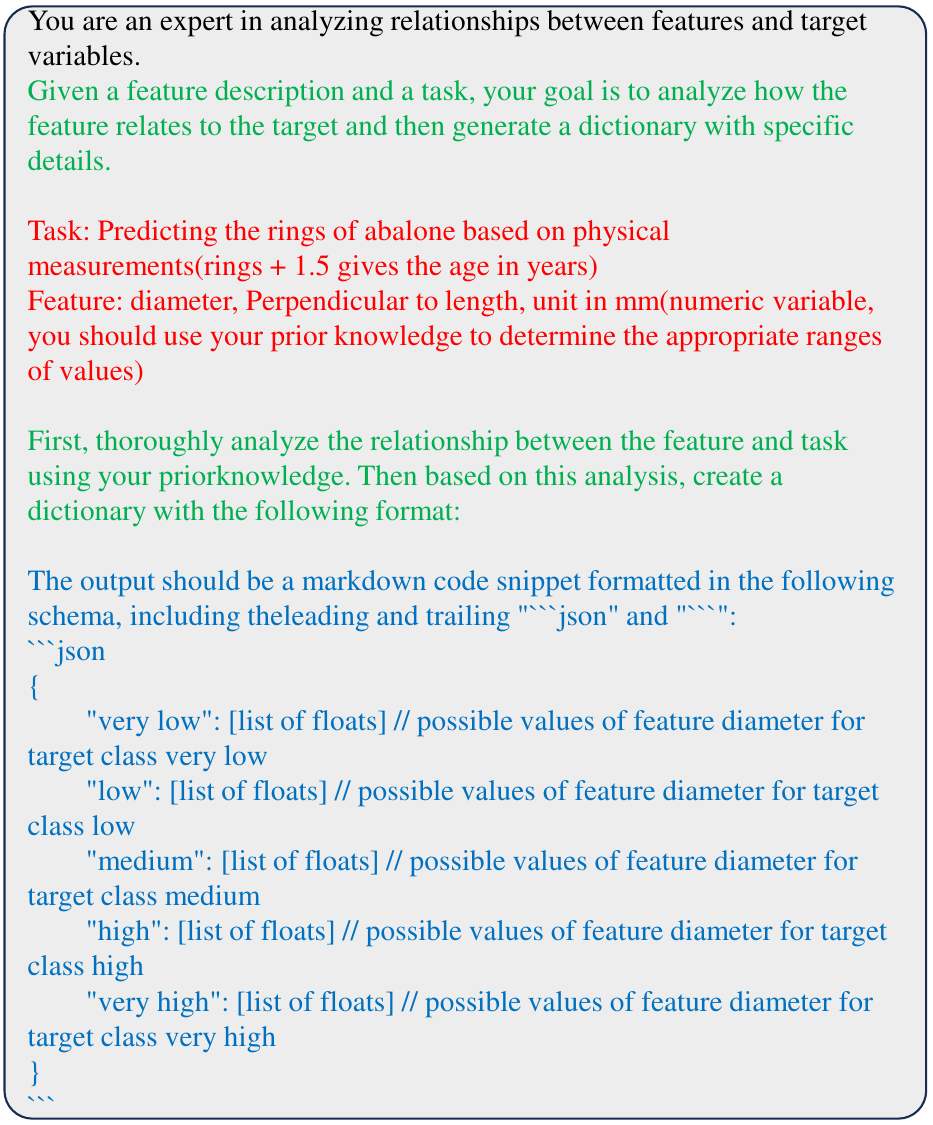}}
    \label{fig:prompt}
    \quad
    \subfloat[answer]{\includegraphics[width=0.99\columnwidth]{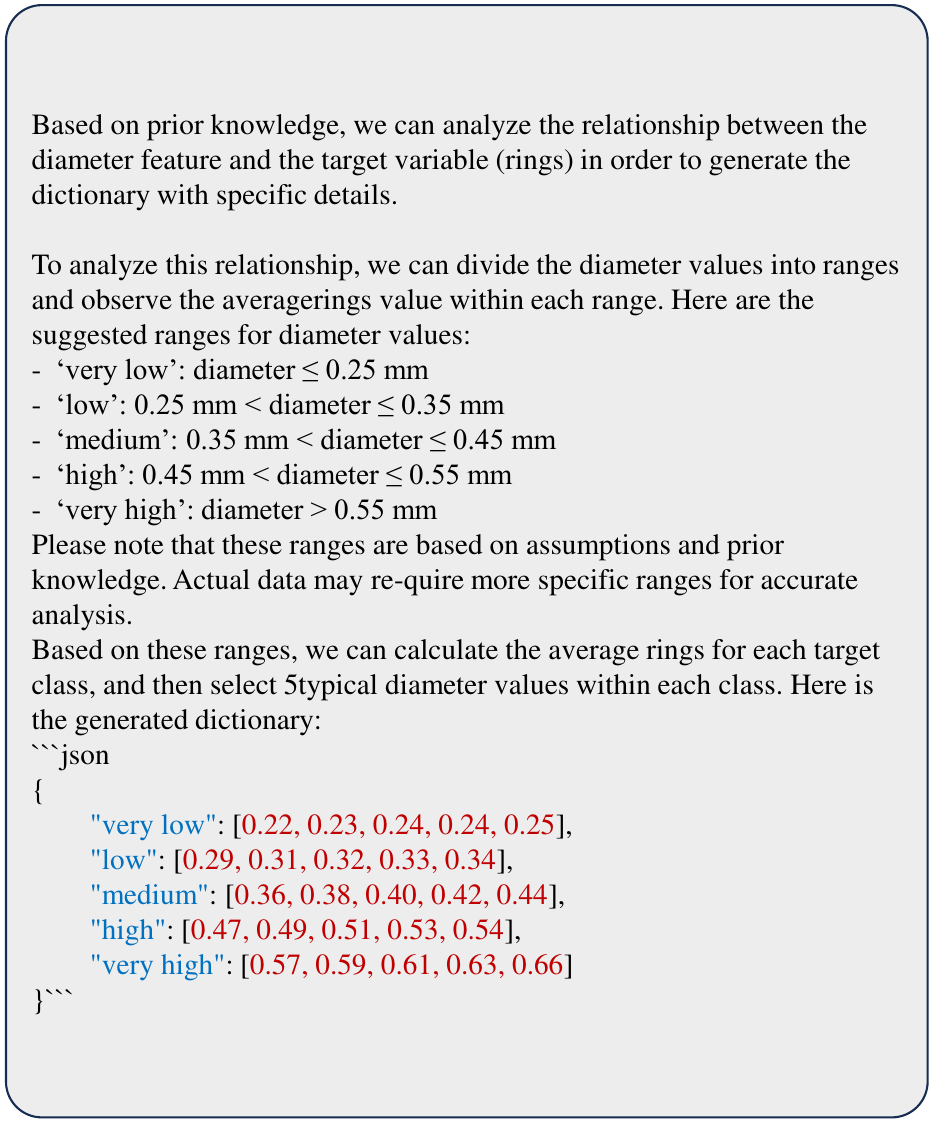}}
    \label{fig:answer}
    
    \caption{\small{{(a) is an example prompt of ProtoLLM for generating feature values about the  ``diameter'' attribute in Abalone dataset. Here, black and red words mean the classic sentence in prompt and descriptions about task and feature, i.e., \textit{\textless Meta-Info\textgreater}. Besides, green words are the reasoning instruction and blue words denote the expected response format, which construct the \textit{\textless Query\textgreater}. (b) is the corresponding response generated by GPT-3.5 for the prompt, where text in blue color denotes the target five classes. Besides, the red words mean the feature values generated by LLMs.}}}
    \label{fig:prompt-answer-regression}
\end{figure*}

Our current prompt design for feature value generation is specifically tailored for classification tasks. To adapt it for regression problems, we convert the continuous target variable into a set of discrete categories within the prompt. Specifically, we partition the target into five ordinal classes: \textit{very low}, \textit{low}, \textit{medium}, \textit{high}, and \textit{very high}. The generated feature values are then combined with traditional machine learning models to enhance performance. The procedure for regression tasks consists of the following three steps:

\paragraph{Query feature values for each discrete class.}
Using the LLM prompt, feature values are generated conditioned on each of the five class labels. The prompt design remains consistent with that used for classification tasks. An example prompt is provided in Fig.~\ref{fig:prompt-answer-regression}.
\paragraph{Estimate representative target values for each class.}
We assume that the target variable $y$ follows a Gaussian distribution with the sample mean as its mean and the sample variance as its variance. Specifically, it is expressed as:
\begin{equation}
    y \sim \mathcal{N}(\mu, \sigma^2)
\end{equation}
where
\begin{equation}
    \begin{aligned}
    \mu = \frac{1}{|\mathcal{S}|} \sum_{(\xv_i,y_i) \in \mathcal{S}} y_i, \\
    \sigma^2 = \frac{1}{|\mathcal{S}|} \sum_{(\xv_i,y_i) \in \mathcal{S}} (y_i - \mu)^2
    \end{aligned}
\end{equation}

The continuous target values are then discretized into five ordinal classes: \textit{very low}, \textit{low}, \textit{medium}, \textit{high}, and \textit{very high}, corresponding to the quantiles:
\[
q = \{0.1, 0.3, 0.5, 0.7, 0.9\}
\]

The representative numerical value for each class is computed using the inverse cumulative distribution function (CDF) of the standard normal distribution:
\[
v_c = \mu + \sigma \cdot \Phi^{-1}(q_c)
\]
where \(\Phi^{-1}\) denotes the probit function (the inverse CDF of the standard normal distribution), and \(q_c\) is the quantile associated with class \(c\). These computed values \(v_c\) serve as representative anchors for each class.

\textbf{Combine generated feature values and corresponding target estimates.} The generated feature values and their associated target values are then integrated to augment traditional machine-learning models for regression.
Specifically, let \( z_{c,d} \) denote the generated value for the \(d\)-th feature corresponding to class \(c\). The augmented sample for class \(c\) is then given by:
\[
(\xv_c, y_c) = (\left[ z_{c,1}, z_{c,2}, \dots, z_{c,D} \right],v_c) 
\]

These augmented samples are subsequently used to train downstream regression models such as MLP, Linear Regression, and K-Nearest Neighbors Regressor, enabling improved performance in few-shot regression scenarios. We report the results in Tab.~\ref{exp:regression}.
\begin{table*}[h]
    \centering
    
    \caption{RMSE results on regression task (lower is better), where we ask LLM to generate 'very low', 'low', 'medium','high','very high' feature values.}
    \label{exp:regression}
    \resizebox{\linewidth}{!}{
\begin{tabular}{c|c|cc|cc|cc}
\toprule
\multirow{2}[4]{*}{dataset} & \multirow{2}[4]{*}{shot} & \multicolumn{2}{c|}{KnnRegressor} & \multicolumn{2}{c|}{LinearRegressor} & \multicolumn{2}{c}{MLP} \\
\cmidrule{3-8}      &       & w/o augmentation & w/ augmentation & w/o augmentation & w/ augmentation & w/o augmentation & w/ augmentation \\
\midrule
\multirow{3}[2]{*}{abalone} & 4     & 3.68±0.89 & \textbf{3.59±0.47} & 10.47±2.82 & \textbf{8.58±2.40} & 10.27±2.56 & \textbf{9.37±1.90} \\
      & 8     & \textbf{3.40±0.28} & \textbf{3.40±0.42} & 13.32±4.86 & \textbf{8.60±1.29} & 10.07±1.32 & \textbf{9.66±1.16} \\
      & 16    & 3.30±0.16 & \textbf{3.13±0.21} & 10.20±0.92 & \textbf{8.89±0.68} & 9.96±0.73 & \textbf{9.79±0.86} \\
\midrule
\multirow{3}[2]{*}{california} & 4     & 127695.29±11599.62 & \textbf{124478.21±11301.49} & 145197.65±67333.22 & \textbf{121296.77±12312.10} & 175730.68±26937.88 & \textbf{168763.81±12040.71} \\
      & 8     & 122038.60±8921.63 & \textbf{117945.30±8558.52} & 164740.07±72540.24 & \textbf{115216.90±10110.49 } & \textbf{165226.72±9595.13} & 168103.52±12025.09 \\
      & 16    & 119206.11±5340.22 & \textbf{115849.27±6541.82} & 156167.49±58812.53 & \textbf{111100.62±9866.29} & \multicolumn{1}{l}{\textbf{164411.93±12080.64}} & 169072.75±8206.45 \\
\midrule
\multirow{3}[2]{*}{CCS} & 4     & 17.57±1.54 & \textbf{16.49±2.16} & 49.79±11.76 & \textbf{32.98±6.93} & 48.68±15.01 & \textbf{36.52±5.92} \\
      & 8     & 17.06±1.26 & \textbf{15.40±1.64} & 72.88±38.55 & \textbf{34.05±5.17} & 48.18±15.50 & \textbf{37.42±4.93} \\
      & 16    & 16.69±1.04 & \textbf{14.64±1.07} & 50.44±18.14 & \textbf{35.78±3.24} & 41.98±4.95 & \textbf{38.68±2.86} \\
\bottomrule
\end{tabular}%
}
    \label{tab:regression}
\end{table*}

\subsubsection{Tasks with More Classes.}
\label{appendix: complex tasks}
We selected several complex datasets to test our method. The Arrhythmia and Soybean datasets each contain more than 10 classes, with the Arrhythmia dataset having over 200 features. These characteristics pose challenges for LLM-based methods due to token limitations, and similarly, other baselines such as TabPFN\_v2 fail to perform effectively. However, our method is able to handle these challenges, as shown in Tab.~\ref{tab:multiclass}. Since both the Arrhythmia and Soybean datasets fail to run on FeatLLM and TabPFN, we also chose the simpler Cmc dataset for comparison.

\begin{table*}[h]
    \centering
    \caption{Comparison of different methods on multiple classification task.}
    \resizebox{0.9\linewidth}{!}{
    \begin{tabular}{c|c|ccccc}
    \toprule
    \textbf{dataset} & \textbf{number of samples per class} & \textbf{LogReg} & \multicolumn{1}{c}{\textbf{KNN}} & \textbf{TabPFN\_v2} & \textbf{FeatLLM} & \textbf{ProtoLLM} \\
    \midrule
    \multirow{2}[2]{*}{\textbf{Arrhythmia}} & 1     & 72.16±3.73 & 52.48±0.00 & N/A   & N/A   & \textbf{72.49±5.24*} \\
          & 2     & 77.70±3.11 & 67.01±2.19 & N/A   & N/A   & \textbf{78.11±4.33*} \\
    \midrule
    \multirow{3}[2]{*}{\textbf{Soybean}} & 1     & \textbf{94.31±1.84} & 52.13±4.18 & N/A   & N/A   & 93.14±2.54 \\
          & 2     & 96.29±2.12 & 52.04±4.32 & N/A   & N/A   & \textbf{96.47±1.37} \\
          & 4     & \textbf{98.25±0.49} & 56.48±3.35 & N/A   & N/A   & 97.82±0.53 \\
    \midrule
    \multirow{3}[2]{*}{\textbf{Cmc}} & 1     & 52.48±6.77 & 52.13±4.18 & 54.21±5.84 & 53.88±6.09 & \textbf{56.62±3.30} \\
          & 2     & 54.14±5.47 & 52.04±4.32 & 54.88±5.16 & 56.37±6.03 & \textbf{57.07±3.15} \\
          & 4     & 57.04±4.36 & 56.48±3.35 & \textbf{59.28±3.01} & 56.01±5.88 & 58.33±3.16 \\
    \bottomrule
    \end{tabular}%
    }
    \label{tab:multiclass}
\end{table*}

\subsubsection{Dataset Information}
The basic information is summarized in Tab.~\ref{tab:addlabel}, and each dataset is introduced as follows:

\begin{itemize} 
\item The \textbf{Abalone}\footnote{\small{\href{https://www.openml.org/search?type=data&status=active&id=44956}{Abalone}}} dataset predicts the age of abalones based on physical measurements such as sex, shell size, and weight. The target variable is the age in years, determined by counting the rings on the shell. 
\item The \textbf{California Housing}\footnote{\small{\href{https://www.openml.org/search?type=data&status=active&id=44025}{California Housing}}} dataset predicts housing prices in California based on features like location, number of rooms, and population. 

\item The \textbf{Concrete Compressive Strength (CCS)}\footnote{\small{\href{https://archive.ics.uci.edu/dataset/165/concrete+compressive+strength}{CCS}}} dataset contains 8 continuous attributes representing concrete ingredients (e.g., cement, blast furnace slag, fly ash, water, superplasticizer, coarse and fine aggregates) and age in days. The target variable is the concrete compressive strength measured in MPa. 
\item The \textbf{Arrhythmia}\footnote{\small{\href{https://www.openml.org/search?type=data&status=active&id=46333}{Arrhythmia}}} dataset aims to classify arrhythmias from ECG recordings, with 279 attributes—206 continuous and the remaining nominal. It categorizes ECGs into 16 classes, distinguishing normal rhythms from various arrhythmia types, with the goal of improving classification accuracy using machine learning techniques. 
\item The \textbf{Soybean}\footnote{\small{\href{https://www.openml.org/search?type=data&status=active&id=42&sort=runs}{Soybean}}} dataset contains 35 categorical attributes related to environmental conditions, plant health, and symptoms of soybean diseases. The dataset is used to classify these diseases into 19 distinct classes. 
\item The \textbf{Contraceptive Method Choice (CMC)}\footnote{\small{\href{https://www.openml.org/search?type=data&status=active&id=23&sort=runs}{Cmc}}} dataset includes 10 attributes, which encompass demographic and socio-economic information, and is used to predict the contraceptive method choice (no use, long-term, or short-term). 
\end{itemize}
\begin{table*}[htbp]
  \centering
  \caption{Dataset information}
  \resizebox{0.5\linewidth}{!}{
    \begin{tabular}{c|cccccc}
    \toprule
          & \multicolumn{1}{l}{Abalone} & \multicolumn{1}{l}{California} & \multicolumn{1}{l}{Arrhythmia} & \multicolumn{1}{l}{Soybean} & \multicolumn{1}{l}{Cmc} \\
    \midrule
    Objects & 4177  & 20640  & 1030 & 452   & 683   & 1473\\
    Numerical & 7     & 8  & 8   & 206   & 0     & 2 \\
    Categorical & 1     & 1   & 0   & 73    & 35    & 7\\
    Class &       &       & & 16    & 19    & 3 \\
    \bottomrule
    \end{tabular}%
    }
  \label{tab:addlabel}%
\end{table*}%

\subsection{Baseline Details}
\label{appendix: baseline details}
In this section, we provide a detailed configuration of the experiments conducted.

For data preprocessing, we apply z-score normalization to numerical features and one-hot encoding to categorical features. In data sampling, the term shot refers to the total number of instances across all classes unless otherwise specified, consistent with the definition in \cite{hegselmann2023tabllm}. For each random seed, 20\% of the dataset is allocated as the test set. The remaining data are balanced by sampling $K$ instances per class, following the approach described in \cite{han2024featllm}. To ensure a fair comparison, columns representing more than 20\% of the dataset are removed, as outlined in the same work.

We adopt the in-context learning variant for LIFT. For P2T, 30 unlabeled samples are selected to extract transferable knowledge; however, this number is reduced to 10 for the credit-g dataset due to token length constraints. During classification, we obtain the log-probabilities of the ten most probable first tokens via the OpenAI API and subsequently assign a probability to each class based on the log-probabilities of the category-related tokens identified within these top ten candidates.

The baseline models of KNN, LogReg, and MLP are implemented using the scikit-learn library,  while XGBoost is implemented using its corresponding package. we utilize the number of samples in a class to determine the parameter \texttt{n\_neighbors} for KNN while utilizing Euclidean distance (with \texttt{p=2}) for distance computation. For LogReg, MLP, and XGboost, we employ a four-fold cross-validation approach along with grid search to identify the optimal hyperparameters, if the sample size is sufficient. The hyperparameter space for LogReg is presented in Tab.~\ref{tab:logreg}. For the MLP, we configure \texttt{early\_stopping} to True, set \texttt{n\_iter\_no\_change} to 5, use \texttt{'adam'} as the solver, specify \texttt{hidden\_layer\_size} as 1024, and limit \texttt{max\_iter} to 200. The other hyperparameter space for MLP is detailed in Tab.~\ref{tab:mlp}. For XGBoost, hyperparameter space of XGBoost is shown in Tab.~\ref{tab:xgb}. Considering TabPFN\_v2, we use the official GitHub repository with the default parameters.
\begin{table}[h!]
    \centering
    \caption{Hyperparameter search space for LogReg.}
    \label{tab:logreg}
        \resizebox{\linewidth}{!}{
            \begin{tabular}{cc}
            \toprule
             Parameter&Search space  \\
             \midrule
             penalty& \{l1, l2\} \\
             \midrule
             C&100, \{10, 1, 1e-1, 1e-2, 1e-3, 1e-4, 1e-5\} \\
             \bottomrule
            \end{tabular}
        }
    
\end{table}

\begin{table}[h!]
    \centering
    \caption{Hyperparameter search space for MLP.}
    \label{tab:mlp}
        \resizebox{\linewidth}{!}{
            \begin{tabular}{cc}
            \toprule
             Parameter&Search space  \\
             \midrule
             alpha& \{1e-3,5e-3,1e-2\} \\
             \midrule
             learning\_rate\_init & \{1e-4,5e-4,1e-3,5e-3,1e-2\} \\
             \bottomrule
            \end{tabular}
        }
    
\end{table}

\begin{table}[h!]
    \centering
    \caption{Hyperparameter search space for XGBoost.}
    \label{tab:xgb}
        \resizebox{\linewidth}{!}{
            \begin{tabular}{cc}
            \toprule
             Parameter&Search space  \\
             \midrule
             max depth& \{2, 4, 6, 8, 10, 12\} \\
             \midrule
             alpha & \{1e-8, 1e-7, 1e-6, 1e-5, 1e-4, 1e-3, 1e-2, 1e-1, 1\} \\
             \midrule
             lambda & \{1e-8, 1e-7, 1e-6, 1e-5, 1e-4, 1e-3, 1e-2, 1e-1, 1\} \\
             eta & \{0.01, 0.03, 0.1, 0.3\} \\
             \bottomrule
            \end{tabular}
        }
    
\end{table}

\subsubsection{ProtoLLM with the T0 backbone.}
\label{appendix:ProtoLLM_T0}
Since TabLLM uses T0 as the base model, we need to exclude model-specific factors when making comparisons. For this reason, we selected T0 for adaptation. To customize our approach for T0, we designed slightly modified prompts compared to those used in our original paper, as T0 is not capable of generating code. We provide a list of possible feature values and allow the T0 model to make selections. For numerical features, the list of values is derived from a few-shot sample. The prompt template is as follows:

\textbf{Template.}
\begin{figure}[htbp]
    \centering
\begin{tikzpicture}
    \node[draw, rounded corners=5pt, text width=\linewidth, align=left] (box) {
Answer choices:\{answer choices\}
\vspace{1em}

\{task for a class\}. Given a list of \{feature name\}(\{feature description\}): \{feature value list\}. Which {feature name} should this belong to?
\vspace{1em}

Answer:
\vspace{1em}
};
\end{tikzpicture}
\caption{Prompt template for the T0 model.}
\end{figure}

\textbf{Example.}
\begin{figure}[htbp]
    \centering
\begin{tikzpicture}
    \node[draw, rounded corners=5pt, text width=\linewidth, align=left] (box) {
Answer choices:Own-child\texttt{|||}Husband\texttt{|||}Not-in-family\texttt{|||}Unmarried\texttt{|||}Wife\texttt{|||}Other-relative.
\vspace{1em}

The person earns more than 50000 dollars per year. Given a list of relationship (what this individual is relative to others):Own-child, Husband, Not-in-family, Unmarried, Wife and Other-relative. Which relationship should this belong to?
\vspace{1em}

Answer:
\vspace{1em}
};
\end{tikzpicture}

\caption{Example prompt for querying feature values of `relationship' for the class yes (indicating the person earns more than 50,000 dollars per year) in the Adult dataset.}
\end{figure}

For numerical features, we sample 10 random values within the range of the feature values from the given few-shot samples to serve as answer choices for the LLM.

As T0 is not well-suited for handling complex tasks such as generating feature weights, we report the results without incorporating feature weights in this experiment. The detailed results in Tab.~\ref{tab:protollm_T0} show that the T0 model can be applied within our framework, although its performance is lower than that of GPT-3.5, which is an expected outcome given the model differences.

\begin{table*}[htbp]
    \centering
    \caption{Results on TabLLM and ProtoLLM with T0 as the Backbone.}
    \resizebox{0.8\linewidth}{!}{
\begin{tabular}{c|c|ccccc}
\toprule
\multirow{2}[4]{*}{dataset} & \multirow{2}[4]{*}{Model} & \multicolumn{5}{c}{shot} \\
\cmidrule{3-7}      &       & 4     & 8     & 16    & 32    & 64 \\
\midrule
\multirow{2}[2]{*}{Adult} & GPT3.5 & 86.01±0.78 & 86.12±0.92 & 86.28±0.77 & 86.26±0.71 & 86.32±0.85 \\
      & T0    & 69.55±4.42 & 76.66±3.63 & 80.81±4.59 & 83.61±1.34 & 85.42±0.81 \\
\midrule
\multirow{2}[2]{*}{Bank} & GPT3.5 & 80.85±2.58 & 81.41±2.58 & 83.26±1.40 & 84.88±1.71 & 85.84±1.28 \\
      & T0    & 54.88±9.40 & 59.92±8.95 & 64.96±6.90 & 69.45±5.84 & 76.24±4.90 \\
\midrule
\multirow{2}[2]{*}{Blood} & GPT3.5 & 75.98±4.99 & 76.35±4.61 & 75.46±4.12 & 75.84±4.39 & 76.08±4.51 \\
      & T0    & 55.37±13.85 & 51.40±18.24 & 56.80±15.67 & 66.79±11.44 & 72.67±9.20 \\
\midrule
\multirow{2}[2]{*}{Car} & GPT3.5 & 79.41±1.92 & 80.40±2.04 & 82.22±2.05 & 84.78±1.81 & 87.45±2.01 \\
      & T0    & 74.64±3.37 & 75.97±3.28 & 78.22±2.98 & 81.56±2.75 & 85.35±2.57 \\
\midrule
\multirow{2}[2]{*}{Credit-g} & GPT3.5 & 62.25±2.86 & 63.26±2.87 & 64.52±3.28 & 68.32±3.05 & 71.75±3.31 \\
      & T0    & 47.01±6.21 & 51.41±7.06 & 55.16±4.67 & 57.00±5.20 & 64.41±4.41 \\
\midrule
\multirow{2}[2]{*}{Diabetes} & GPT3.5 & 75.68±3.61 & 75.76±3.78 & 75.70±4.33 & 77.48±3.81 & 78.00±3.86 \\
      & T0    & 73.31±6.01 & 70.75±8.24 & 74.94±6.03 & 78.18±3.37 & 76.75±4.12 \\
\midrule
\multirow{2}[2]{*}{Heart} & GPT3.5 & 70.42±7.69 & 75.16±8.03 & 81.85±4.50 & 85.69±3.34 & 89.12±1.88 \\
      & T0    & 71.20±12.26 & 79.59±6.93 & 82.17±7.41 & 87.90±2.55 & 89.27±2.01 \\
\bottomrule
\end{tabular}%
    } \label{tab:protollm_T0}
\end{table*}

\subsection{Detailed Results}
\label{appendix:detail_results}
This chapter presents detailed experimental results, including the main results of LLM-based and non-LLM-based methods, as well as various ablation studies.
\begin{table*}[htbp]
\centering  
  \caption{AUC performance across 10 datasets in few-shot scenarios compared with LLM-based method. Results marked with * indicate performance without using feature weights due to the token limit. 'N/A' denotes cases where models could not run due to token length (LLM-based).}
  \label{tab: main_results}
\resizebox{0.85\linewidth}{!}{
\begin{tabular}{c|c|ccccccc}
\toprule
Data  & Shot  & In-context & TABLET & LIFT  & P2T   & TabLLM & FeatLLM & ProtoLLM \\
\midrule
\multirow{5}[2]{*}{Adult} & 4     & 77.51±5.24 & 75.29±12.24 & 83.07±2.29 & 82.66±2.78 & 83.57±2.69 & \textbf{86.68±0.86} & \underline{83.74±1.50} \\
      & 8     & 79.30±2.89 & 77.56±7.56 & 83.58±1.57 & 83.56±3.75 & 83.52±4.30 & \textbf{87.89±0.06} & \underline{83.95±1.61} \\
      & 16    & 79.50±4.57 & 79.74±5.64 & 84.96±3.05 & \underline{86.02±1.96} & 83.23±2.45 & \textbf{87.54±0.50} & 84.00±1.56 \\
      & 32    & 81.89±4.04 & 78.08±6.70 & \underline{84.60±2.74} & 84.11±1.73 & 82.60±4.14 & \textbf{87.09±0.58} & 83.97±1.48 \\
      & 64    & N/A   & N/A   & N/A   & N/A   & \underline{84.88±0.97} & \textbf{87.77±0.31} & 84.25±1.44 \\
\midrule
\multirow{5}[2]{*}{Bank} & 4     & 61.38±1.30 & 58.11±6.29 & \underline{71.90±4.33} & 70.50±3.94 & 62.51±8.95 & 70.45±3.69 & \textbf{82.53±1.73} \\
      & 8     & 69.57±13.35 & 69.08±6.00 & 66.56±6.37 & 66.75±4.95 & 63.19±5.79 & \underline{75.85±6.66} & \textbf{82.71±1.95} \\
      & 16    & 69.76±8.55 & 69.40±11.28 & 69.97±4.59 & 72.02±4.61 & 63.73±6.43 & \underline{78.41±1.08} & \textbf{83.13±1.60} \\
      & 32    & 66.93±5.67 & 73.61±9.28 & 69.29±4.96 & 72.21±4.18 & 66.51±3.92 & \underline{78.37±4.50} & \textbf{83.86±1.24} \\
      & 64    & N/A   & N/A   & N/A   & N/A   & 70.83±3.43 & \underline{81.18±6.17} & \textbf{84.52±0.64} \\
\midrule
\multirow{5}[2]{*}{Blood} & 4     & 56.30±12.43 & 56.45±15.45 & \underline{69.86±6.05} & 63.69±8.63 & 55.87±13.49 & 68.34±7.48 & \textbf{76.31±5.12} \\
      & 8     & 58.99±10.12 & 56.37±11.56 & 64.40±5.73 & 62.52±5.70 & 66.01±9.25 & \underline{70.37±3.23} & \textbf{76.39±4.62} \\
      & 16    & 56.59±5.21 & 60.62±4.13 & 56.71±2.92 & 62.12±7.72 & 65.14±7.55 & \underline{70.07±5.19} & \textbf{75.98±4.52} \\
      & 32    & 58.69±1.53 & 57.94±4.16 & 56.51±3.52 & 67.68±4.95 & 69.95±3.39 & \underline{71.13±4.38} & \textbf{76.16±4.54} \\
      & 64    & 65.79±2.05 & 63.47±7.36 & 58.79±4.67 & 67.94±5.45 & 70.88±1.58 & \underline{71.04±4.36} & \textbf{75.97±4.29} \\
\midrule
\multirow{5}[2]{*}{Car} & 4     & 62.47±2.47 & 60.21±4.81 & 70.16±4.02 & 50.36±5.13 & \textbf{85.82±3.65} & 72.69±1.52 & \underline{78.88±3.12} \\
      & 8     & 67.57±3.44 & 65.53±8.00 & 76.72±3.55 & 54.46±1.94 & \textbf{87.43±2.56} & 73.26±1.46 & \underline{79.67±2.99} \\
      & 16    & 76.94±3.04 & 74.02±1.01 & 77.98±3.82 & 61.00±3.75 & \textbf{88.65±2.63} & 79.43±1.24 & \underline{81.02±2.53} \\
      & 32    & 81.64±2.52 & 76.44±4.02 & 78.34±4.55 & 62.48±4.63 & \textbf{89.02±1.50} & \underline{85.01±1.36} & 82.42±2.22 \\
      & 64    & 77.65±3.74 & 76.13±1.17 & 83.10±5.27 & N/A   & \textbf{92.18±0.47} & \underline{86.78±0.90} & 84.10±2.86 \\
\midrule
\multirow{5}[2]{*}{Credit-g} & 4     & 52.99±4.08 & 54.33±6.54 & 52.89±3.38 & 49.60±5.81 & 51.90±9.40 & \underline{55.94±1.10} & \textbf{61.43±4.19} \\
      & 8     & 52.43±4.36 & 52.90±5.79 & 53.34±3.50 & 50.92±4.60 & 56.42±12.89 & \underline{57.42±3.10} & \textbf{63.26±3.73} \\
      & 16    & 55.29±4.80 & 51.65±4.02 & 51.03±3.65 & 54.05±2.71 & \underline{60.38±14.03} & 56.60±2.22 & \textbf{65.22±3.46} \\
      & 32    & N/A   & N/A   & N/A   & N/A   & \underline{68.64±3.86} & 61.79±10.25 & \textbf{69.03±2.52} \\
      & 64    & N/A   & N/A   & N/A   & N/A   & \underline{70.80±4.09} & 66.43±2.90 & \textbf{72.60±2.53} \\
\midrule
\multirow{5}[2]{*}{Cultivars} & 4     & 51.38±2.48 & 54.28±3.73 & \underline{57.61±8.45} & 51.80±4.69 & 54.39±5.61 & 55.63±5.24 & \textbf{57.73±7.64} \\
      & 8     & 51.68±4.43 & 51.48±3.85 & \underline{58.65±1.99} & 52.58±6.83 & 52.86±6.13 & 56.97±5.08 & \textbf{59.35±8.42} \\
      & 16    & 54.31±6.12 & 57.44±3.53 & \textbf{62.92±4.84} & 53.52±9.57 & 56.97±2.22 & 57.19±5.30 & \underline{60.45±7.80} \\
      & 32    & N/A   & N/A   & 54.70±2.72 & \underline{61.39±6.13} & 58.50±2.65 & 59.62±7.43 & \textbf{62.01±7.31} \\
      & 64    & N/A   & N/A   & N/A   & N/A   & \underline{60.32±2.60} & 59.14±4.79 & \textbf{65.09±7.40} \\
\midrule
\multirow{5}[2]{*}{Diabetes} & 4     & 71.71±5.31 & 63.96±3.32 & 77.69±2.67 & 73.01±2.98 & 70.42±3.69 & \underline{80.28±0.75} & \textbf{80.45±2.29} \\
      & 8     & 72.21±2.07 & 65.47±3.95 & 77.02±1.41 & 74.95±4.16 & 64.30±5.88 & \underline{79.38±1.66} & \textbf{80.48±2.26} \\
      & 16    & 71.64±5.05 & 66.71±0.76 & 74.30±2.65 & 73.35±4.12 & 67.34±2.79 & \underline{80.15±1.35} & \textbf{80.46±2.38} \\
      & 32    & 73.32±1.59 & 66.97±1.75 & 74.14±2.99 & 75.94±1.82 & 69.74±4.41 & \underline{80.06±1.18} & \textbf{81.02±2.71} \\
      & 64    & 70.22±4.09 & 69.27±6.15 & N/A   & N/A   & 71.56±4.55 & \underline{80.91±1.62} & \textbf{81.50±2.20} \\
\midrule
\multirow{5}[2]{*}{Heart} & 4     & 60.76±4.00 & 68.19±11.17 & 73.73±3.03 & \underline{74.97±2.25} & 59.74±4.49 & \textbf{75.66±4.59} & 74.12±5.60 \\
      & 8     & 65.46±3.77 & 69.85±10.82 & 76.30±2.88 & 77.01±2.35 & 70.14±7.91 & \textbf{79.46±2.16} & \underline{77.52±4.99} \\
      & 16    & 67.00±7.83 & 68.39±11.73 & 80.43±2.03 & 80.06±1.81 & \underline{81.72±3.92} & \textbf{83.71±1.88} & 81.56±3.57 \\
      & 32    & 71.94±3.88 & 71.90±9.07 & 81.27±2.22 & 82.85±0.82 & \textbf{87.43±2.32} & \underline{87.19±3.66} & 83.94±3.15 \\
      & 64    & N/A   & N/A   & N/A   & N/A   & \textbf{89.78±2.59} & \underline{88.08±4.11} & 86.25±2.24 \\
\midrule
\multirow{5}[2]{*}{Myocardial} & 4     & N/A   & N/A   & N/A   & N/A   & N/A   & \underline{52.87±3.44} & \textbf{63.25±4.16*} \\
      & 8     & N/A   & N/A   & N/A   & N/A   & N/A   & \underline{56.22±1.64} & \textbf{63.62±4.12*} \\
      & 16    & N/A   & N/A   & N/A   & N/A   & N/A   & \underline{55.32±9.15} & \textbf{64.03±4.04*} \\
      & 32    & N/A   & N/A   & N/A   & N/A   & N/A   & \underline{60.02±4.02} & \textbf{65.44±4.38*} \\
      & 64    & N/A   & N/A   & N/A   & N/A   & N/A   & \underline{61.47±3.91} & \textbf{65.75±4.34*} \\
\midrule
\multirow{5}[2]{*}{NHANES} & 4     & 91.84±3.79 & 93.54±4.20 & 90.19±2.21 & 97.96±0.53 & \textbf{99.49±0.23} & 92.20±1.71 & \underline{98.10±1.03} \\
      & 8     & 86.67±5.49 & 94.25±3.35 & 89.89±2.63 & 98.15±0.96 & \textbf{100.00±0.00} & 93.29±7.01 & \underline{98.34±1.14} \\
      & 16    & 93.33±4.47 & 95.02±1.57 & 89.21±2.24 & 97.71±1.23 & \textbf{100.00±0.00} & 95.64±4.67 & \underline{99.09±0.67} \\
      & 32    & 88.54±5.40 & 95.82±3.71 & 86.93±9.91 & 98.73±0.70 & \textbf{100.00±0.00} & 97.29±1.28 & \underline{99.42±0.43} \\
      & 64    & N/A   & N/A   & N/A   & N/A   & \textbf{100.00±0.00} & 98.32±0.65 & \underline{99.65±0.20} \\
\midrule
\multirow{5}[2]{*}{Average AUC} & 4     & N/A   & N/A   & N/A   & N/A   & N/A   & \underline{71.07} & \textbf{75.65} \\
      & 8     & N/A   & N/A   & N/A   & N/A   & N/A   & \underline{73.01} & \textbf{76.53} \\
      & 16    & N/A   & N/A   & N/A   & N/A   & N/A   & \underline{74.41} & \textbf{77.49} \\
      & 32    & N/A   & N/A   & N/A   & N/A   & N/A   & \underline{76.76} & \textbf{78.73} \\
      & 64    & N/A   & N/A   & N/A   & N/A   & N/A   & 78.11 & \textbf{79.97} \\
\midrule
\multirow{5}[2]{*}{Average Rank} & 4     & 5.67  & 5.44  & 3.67  & 4.56  & 4.44  & \underline{2.6} & \textbf{1.5} \\
      & 8     & 5.67  & 5.78  & 3.89  & 4.78  & 4     & \underline{2.4} & \textbf{1.4} \\
      & 16    & 5.78  & 5.56  & 4.33  & 4.44  & 3.56  & \underline{2.4} & \textbf{1.8} \\
      & 32    & 5.43  & 5.86  & 5     & 3.75  & 3.33  & \underline{2.3} & \textbf{1.8} \\
      & 64    & 4.67  & 5.67  & 5.5   & 4     & \underline{2} & 2.2   & \textbf{1.7} \\
\bottomrule
\end{tabular}%

}
\end{table*}

\begin{table*}[]
    \centering
     \caption{AUC performance across 10 datasets in few-shot scenarios compared with non-LLM-based methods. Results marked with * indicate performance without using feature weights due to the token limit.}
     \label{tab:non-LLM-based methods results}
    \resizebox{0.85\linewidth}{!}{
\begin{tabular}{c|c|ccccccc}
\toprule
Dataset & shot  & XGB   & KNN   & MLP   & STUNT & LogReg & TabPFN\_v2 & ProtoLLM \\
\midrule
\multirow{5}[2]{*}{Adult} & 4     & 50.00±0.00 & 61.39±8.23 & 55.16±13.97 & 67.43±29.61 & 65.52±12.63 & \underline{68.46±10.25} & \textbf{83.74±1.50} \\
      & 8     & 57.68±6.80 & 72.43±4.79 & 71.24±8.13 & \underline{82.16±6.93} & 71.90±9.16 & 74.95±6.45 & \textbf{83.95±1.61} \\
      & 16    & 72.96±4.79 & 78.25±3.01 & 78.01±9.19 & \underline{80.57±10.93} & 78.27±7.46 & 79.11±3.41 & \textbf{84.00±1.56} \\
      & 32    & 76.02±3.39 & 81.59±2.28 & 80.91±6.56 & 78.08±15.15 & 81.82±5.23 & \underline{81.84±2.76} & \textbf{83.97±1.48} \\
      & 64    & 80.24±2.77 & 84.14±1.32 & \underline{85.39±1.90} & \textbf{86.01±0.16} & 84.54±2.97 & 84.97±1.58 & 84.25±1.44 \\
\midrule
\multirow{5}[2]{*}{Bank} & 4     & 50.00±0.00 & 57.96±4.82 & 56.87±8.76 & 56.34±12.82 & 59.29±9.86 & \underline{64.23±5.48} & \textbf{82.53±1.73} \\
      & 8     & 56.05±9.29 & 63.13±5.90 & 61.86±10.23 & 63.01±8.78 & 66.46±12.23 & \underline{69.43±7.50} & \textbf{82.71±1.95} \\
      & 16    & 69.86±7.85 & 69.38±4.44 & 66.80±10.31 & 69.85±0.95 & 74.15±6.95 & \underline{78.48±5.56} & \textbf{83.13±1.60} \\
      & 32    & 73.99±3.94 & 73.43±4.80 & 71.99±7.52 & 71.64±1.65 & 78.25±4.29 & \underline{80.47±4.79} & \textbf{83.86±1.24} \\
      & 64    & 79.53±3.23 & 77.80±3.34 & 79.35±4.30 & 72.26±1.62 & 81.61±3.19 & \underline{83.35±2.98} & \textbf{84.52±0.64} \\
\midrule
\multirow{5}[2]{*}{Blood} & 4     & 50.00±0.00 & 53.33±7.87 & 54.90±16.37 & 48.57±6.04 & \underline{58.02±13.35} & 57.63±14.91 & \textbf{76.31±5.12} \\
      & 8     & 52.79±8.66 & 57.73±6.13 & \underline{63.40±9.44} & 60.00±4.84 & 57.20±11.28 & 59.30±8.79 & \textbf{76.39±4.62} \\
      & 16    & 60.55±9.23 & 64.68±10.04 & \underline{65.92±11.80} & 54.76±4.53 & 65.41±11.70 & 65.08±8.25 & \textbf{75.98±4.52} \\
      & 32    & 65.54±8.31 & 68.55±7.22 & 62.82±15.64 & 59.87±3.72 & \underline{72.30±9.21} & 69.84±8.14 & \textbf{76.16±4.54} \\
      & 64    & 68.67±5.44 & 72.46±5.33 & 72.72±7.34 & 61.75±2.19 & \underline{74.86±6.68} & 73.64±5.39 & \textbf{75.97±4.29} \\
\midrule
\multirow{5}[2]{*}{Car} & 4     & 50.00±0.00 & 60.14±4.74 & 52.85±5.47 & 61.32±3.83 & \underline{65.14±5.53} & 62.80±5.23 & \textbf{78.88±3.12} \\
      & 8     & 60.48±4.56 & 65.02±2.93 & 62.90±8.01 & 67.86±0.49 & 65.05±7.13 & \underline{67.94±4.21} & \textbf{79.67±2.99} \\
      & 16    & 70.07±5.35 & 72.23±2.12 & 75.34±5.22 & 75.56±2.88 & 76.33±2.45 & \underline{79.72±5.39} & \textbf{81.02±2.53} \\
      & 32    & 80.06±3.53 & 77.50±2.94 & 83.56±3.11 & 82.29±2.34 & \underline{84.95±2.61} & \textbf{89.84±3.03} & 82.42±2.22 \\
      & 64    & 89.50±2.82 & 81.32±2.25 & 87.86±3.73 & 84.45±1.69 & \underline{91.69±2.49} & \textbf{95.77±1.80} & 84.10±2.86 \\
\midrule
\multirow{5}[2]{*}{Credit-g} & 4     & 50.00±0.00 & 53.33±3.87 & 51.27±6.47 & 48.80±6.76 & \underline{54.01±5.42} & 54.73±7.88 & \textbf{61.43±4.19} \\
      & 8     & 55.71±4.87 & 54.26±4.91 & 53.60±7.88 & 54.50±8.25 & 58.15±7.63 & \underline{59.36±5.80} & \textbf{63.26±3.73} \\
      & 16    & 59.28±5.04 & 56.89±5.62 & 55.39±7.83 & 57.63±7.58 & 58.62±7.92 & \underline{61.62±5.55} & \textbf{65.22±3.46} \\
      & 32    & 65.26±4.49 & 61.06±3.53 & 60.50±7.59 & 63.24±5.47 & 64.16±5.27 & \underline{68.34±3.55} & \textbf{69.03±2.52} \\
      & 64    & 68.12±3.51 & 66.06±3.70 & 65.96±8.45 & \underline{70.97±4.95} & 68.51±5.27 & 69.39±4.41 & \textbf{72.60±2.53} \\
\midrule
\multirow{5}[2]{*}{Cultivars} & 4     & 50.00±0.00 & 45.84±7.29 & 43.95±7.55 & \underline{57.10±8.66} & 44.98±7.89 & 45.36±5.56 & \textbf{57.73±7.64} \\
      & 8     & 50.86±8.09 & 47.47±8.93 & 47.64±10.07 & \underline{57.26±9.52} & 50.20±8.63 & 46.94±7.55 & \textbf{59.35±8.42} \\
      & 16    & 48.48±9.09 & 47.62±9.58 & 48.86±10.01 & \underline{60.09±7.64} & 48.48±7.76 & 48.86±7.84 & \textbf{60.45±7.80} \\
      & 32    & 52.96±7.40 & 50.54±10.05 & 57.20±10.26 & \underline{60.48±6.51} & 53.15±8.63 & 52.12±9.50 & \textbf{62.01±7.31} \\
      & 64    & 57.41±8.20 & 51.01±10.37 & \textbf{67.51±6.43} & 61.07±6.77 & 63.70±9.65 & 55.89±7.64 & \underline{65.09±7.40} \\
\midrule
\multirow{5}[2]{*}{Diabetes} & 4     & 50.00±0.00 & 59.48±6.89 & 58.53±14.21 & 64.22±6.78 & 58.74±13.20 & \underline{65.09±11.57} & \textbf{80.45±2.29} \\
      & 8     & 59.30±12.01 & 63.70±8.14 & 63.73±8.25 & 67.39±12.92 & 70.79±5.87 & \underline{71.57±8.43} & \textbf{80.48±2.26} \\
      & 16    & 66.88±8.42 & 68.30±5.73 & 66.61±7.82 & 73.79±6.48 & 66.34±8.61 & \underline{73.59±7.32} & \textbf{80.46±2.38} \\
      & 32    & 72.30±4.17 & 73.96±4.70 & 74.39±4.73 & 76.70±4.55 & 77.44±5.13 & \underline{78.67±5.01} & \textbf{81.02±2.71} \\
      & 64    & 74.40±4.10 & 77.59±2.84 & 77.00±6.18 & 78.64±3.32 & 78.72±3.57 & \underline{79.09±4.21} & \textbf{81.50±2.20} \\
\midrule
\multirow{5}[2]{*}{Heart} & 4     & 50.00±0.00 & 63.98±11.68 & 61.15±18.38 & \textbf{88.27±3.32} & 63.98±19.36 & \underline{75.15±18.95} & 74.12±5.60 \\
      & 8     & 59.00±12.07 & 73.64±11.23 & 77.36±10.24 & \textbf{88.78±2.38} & 76.93±10.21 & \underline{83.96±7.27} & 77.52±4.99 \\
      & 16    & 83.61±5.06 & 84.23±4.08 & 84.91±7.70 & \textbf{89.13±2.10} & 85.27±4.82 & \underline{87.51±3.69} & 81.56±3.57 \\
      & 32    & 85.81±3.83 & 87.88±2.22 & 87.33±5.08 & \textbf{89.65±3.04} & 88.74±2.72 & \underline{89.46±2.50} & 83.94±3.15 \\
      & 64    & 87.21±2.97 & 88.91±1.69 & \underline{90.17±1.85} & 89.62±3.16 & 89.50±2.19 & \textbf{90.21±2.39} & 86.25±2.24 \\
\midrule
\multirow{5}[2]{*}{Myocardial} & 4     & 50.00±0.00 & 53.28±5.65 & 53.87±9.22 & 52.77±2.01 & 54.88±8.06 & \underline{54.95±7.36} & \textbf{63.25±4.16*} \\
      & 8     & 52.65±5.67 & 55.92±6.44 & 55.25±7.53 & 55.40±4.41 & \underline{56.36±6.18} & 54.29±7.63 & \textbf{63.62±4.12*} \\
      & 16    & 52.75±7.37 & 54.49±6.75 & 55.21±6.93 & \underline{61.22±3.45} & 54.77±5.87 & 59.06±6.85 & \underline{64.03±4.04*} \\
      & 32    & 58.78±8.04 & 58.88±5.03 & 57.21±7.91 & 60.76±1.58 & 63.03±6.53 & \textbf{65.96±7.26} & \underline{65.44±4.38*} \\
      & 64    & 63.03±7.84 & 60.04±3.67 & 58.81±8.67 & 59.79±0.56 & 64.49±5.90 & \textbf{66.57±7.85} & \underline{65.75±4.34*} \\
\midrule
\multirow{5}[2]{*}{NHANES} & 4     & 50.00±0.00 & 64.90±8.82 & 59.58±14.23 & 69.32±19.59 & 77.56±15.27 & \underline{87.64±10.55} & \textbf{98.10±1.03} \\
      & 8     & 91.93±6.99 & 75.38±5.89 & 76.43±6.86 & 68.56±18.35 & 88.34±12.64 & \underline{97.43±3.22} & \textbf{98.34±1.14} \\
      & 16    & 94.25±6.09 & 86.34±4.71 & 86.34±6.05 & 68.62±19.81 & 97.12±4.35 & \textbf{99.95±0.08} & \textbf{99.09±0.67} \\
      & 32    & 98.31±2.88 & 92.18±2.55 & 91.88±4.09 & 75.06±3.56 & 99.30±0.56 & \textbf{100.00±0.00} & \underline{99.42±0.43} \\
      & 64    & 99.87±0.47 & 94.79±2.16 & 95.32±2.86 & 80.29±4.56 & 99.49±0.89 & \textbf{100.00±0.00} & \underline{99.65±0.20} \\
\midrule
\multirow{5}[2]{*}{Average AUC} & 4     & 50    & 57.363 & 54.813 & 61.414 & 60.212 & \underline{63.604} & \textbf{75.654} \\
      & 8     & 59.645 & 62.868 & 63.341 & 66.492 & 66.138 & \underline{68.517} & \textbf{76.529} \\
      & 16    & 67.869 & 68.241 & 68.339 & 69.122 & 70.476 & \underline{73.298} & \textbf{77.494} \\
      & 32    & 72.903 & 72.557 & 72.779 & 71.777 & 76.314 & \underline{77.654} & \textbf{78.727} \\
      & 64    & 76.798 & 75.412 & 78.009 & 74.485 & 79.711 & \underline{79.888} & \textbf{79.968} \\
\bottomrule
\end{tabular}%

    }

\end{table*}

\section{results}
\begin{table*}[htbp]
  \centering
  \caption{AUC across 9 datasets with different generation types. \texttt{F} denotes feature-level generation and \texttt{D} denotes decomposition into sub-problems. \texttt{E} denotes example-based prompts for LLMs.}
  \label{tab: results of different generation types}
  \resizebox{\linewidth}{!}{
\begin{tabular}{c|c|cclccccc}
\toprule
      &       & E   & E-W & \multicolumn{1}{c}{E-D} & E-D-W & S     & F     & W   & D-W \\
\midrule
\multirow{4}[2]{*}{Adult} & 0     &     N/A  &  N/A     &  N/A      &   N/A    & 83.69±2.36 & 85.93±0.64 & 81.83±3.66 & 83.72±1.41 \\
      & 4     & 79.41±4.09 & 77.60±4.02 & 79.44±4.98 & 81.52±0.98 & 79.34±4.97 & 86.01±0.78 & 77.84±3.50 & 83.74±1.50 \\
      & 8     & 81.63±2.57 & 80.22±3.58 & 83.47±1.52 & 82.76±2.38 & 82.41±2.43 & 86.12±0.92 & 81.05±3.11 & 83.95±1.61 \\
      & 16    & 82.43±3.07 & 81.20±2.95 & 84.26±2.95 & 82.09±3.19 & 82.70±3.16 & 86.28±0.77 & 81.27±3.48 & 84.00±1.56 \\
\midrule
\multirow{4}[2]{*}{Bank} & 0     &     N/A  &  N/A     &  N/A      &   N/A    & 72.56±5.88 & 80.20±2.22 & 74.14±4.92 & 82.93±1.90 \\
      & 4     & 73.80±11.61 & 72.00±10.22 & 70.28±6.72 & 74.11±6.43 & 74.98±11.41 & 80.85±2.58 & 75.03±9.13 & 82.53±1.73 \\
      & 8     & 75.65±11.46 & 75.99±8.63 & 71.01±5.62 & 74.50±7.63 & 75.75±10.75 & 81.41±2.58 & 77.02±7.86 & 82.71±1.95 \\
      & 16    & 80.40±2.41 & 79.81±3.21 & 76.85±3.82 & 79.29±2.49 & 80.49±1.98 & 83.26±1.40 & 81.14±2.43 & 83.13±1.60 \\
\midrule
\multirow{4}[2]{*}{Blood} & 0     &     N/A  &  N/A     &  N/A      &   N/A    & 70.31±5.52 & 75.63±4.15 & 73.52±4.95 & 75.31±4.28 \\
      & 4     & 58.53±14.48 & 62.41±15.36 & 62.93±12.96 & 65.04±11.77 & 62.24±14.80 & 75.98±4.99 & 65.21±14.72 & 76.31±5.12 \\
      & 8     & 61.41±10.51 & 66.75±7.60 & 64.39±10.31 & 66.19±9.40 & 63.75±10.04 & 76.35±4.61 & 69.31±7.61 & 76.39±4.62 \\
      & 16    & 68.49±10.69 & 70.41±9.52 & 68.46±10.02 & 69.68±10.57 & 68.96±9.62 & 75.46±4.12 & 71.88±8.82 & 75.98±4.52 \\
\midrule
\multirow{4}[2]{*}{Car} & 0     &     N/A  &  N/A     &  N/A      &   N/A    & 68.97±4.45 & 78.29±1.72 & 69.39±4.50 & 77.76±2.29 \\
      & 4     & 72.92±5.04 & 72.01±5.04 & 68.84±5.22 & 69.57±6.43 & 73.50±5.46 & 79.41±1.92 & 73.00±5.00 & 78.88±3.12 \\
      & 8     & 75.92±3.73 & 74.39±4.73 & 72.69±2.96 & 72.98±6.17 & 76.86±4.14 & 80.40±2.04 & 76.10±5.01 & 79.67±2.99 \\
      & 16    & 79.64±3.53 & 78.31±3.62 & 79.37±3.02 & 80.28±2.34 & 80.61±3.33 & 82.22±2.05 & 79.36±2.95 & 81.02±2.53 \\
\midrule
\multirow{4}[2]{*}{Credit-g} & 0     &     N/A  &  N/A     &  N/A      &   N/A    & 56.34±4.68 & 61.29±3.03 & 54.11±4.63 & 59.64±3.15 \\
      & 4     & 56.24±8.18 & 56.86±7.47 & 58.11±4.92 & 57.75±5.82 & 59.15±5.44 & 62.25±2.86 & 58.26±6.93 & 61.43±4.19 \\
      & 8     & 60.42±4.58 & 59.84±6.39 & 60.71±3.42 & 61.69±4.37 & 62.40±4.47 & 63.26±2.87 & 61.58±5.63 & 63.26±3.73 \\
      & 16    & 63.21±7.15 & 62.00±7.33 & 60.57±6.10 & 60.82±6.64 & 64.33±7.03 & 64.52±3.28 & 63.29±6.98 & 65.22±3.46 \\
\midrule
\multirow{4}[2]{*}{Cultivars} & 0     &     N/A  &  N/A     &  N/A      &   N/A    & 50.66±8.20 & 58.93±8.13 & 51.62±8.07 & 59.12±7.46 \\
      & 4     & 43.50±7.57 & 43.27±7.96 & 47.71±9.54 & 47.47±8.85 & 45.79±7.48 & 59.37±7.98 & 44.73±8.95 & 57.73±7.64 \\
      & 8     & 48.48±9.43 & 47.78±9.04 & 49.62±9.96 & 49.65±8.55 & 49.94±9.51 & 60.51±8.00 & 49.22±9.82 & 59.35±8.42 \\
      & 16    & 48.64±8.94 & 47.63±8.03 & 50.07±9.20 & 49.48±8.70 & 53.41±7.52 & 60.45±7.13 & 52.42±8.27 & 60.45±7.80 \\
\midrule
\multirow{4}[2]{*}{Diabetes} & 0     &     N/A  &  N/A     &  N/A      &   N/A    & 82.29±2.36 & 75.65±3.05 & 82.12±2.63 & 80.32±2.03 \\
      & 4     & 77.61±5.32 & 76.73±4.97 & 71.57±6.28 & 76.27±6.09 & 78.24±4.78 & 75.68±3.61 & 77.84±4.81 & 80.45±2.29 \\
      & 8     & 78.22±4.56 & 76.92±5.71 & 71.55±8.03 & 76.14±6.69 & 79.19±3.67 & 75.76±3.78 & 78.21±4.63 & 80.48±2.26 \\
      & 16    & 79.32±3.50 & 79.18±3.08 & 74.48±5.05 & 79.60±3.20 & 79.42±3.43 & 75.70±4.33 & 79.60±3.33 & 80.46±2.38 \\
\midrule
\multirow{4}[2]{*}{Heart} & 0     &     N/A  &  N/A     &  N/A      &   N/A    & 77.86±2.37 & 62.40±4.34 & 74.58±4.07 & 68.51±4.77 \\
      & 4     & 74.11±19.44 & 77.98±11.39 & 70.53±18.59 & 76.30±5.81 & 80.69±11.24 & 70.42±7.69 & 78.28±10.63 & 74.12±5.60 \\
      & 8     & 82.09±9.37 & 82.35±7.01 & 80.79±9.61 & 79.52±5.66 & 85.38±4.28 & 75.16±8.03 & 82.25±6.01 & 77.52±4.99 \\
      & 16    & 87.45±3.77 & 85.50±4.06 & 86.47±3.72 & 82.60±3.68 & 88.21±2.78 & 81.85±4.50 & 85.75±3.67 & 81.56±3.57 \\
\midrule
\multirow{4}[2]{*}{NHANES} & 0     &     N/A  &  N/A     &  N/A      &   N/A    & 96.32±1.50 & 83.96±3.21 & 98.36±1.05 & 97.48±1.20 \\
      & 4     & 95.09±3.79 & 97.07±2.64 & 76.04±12.39 & 96.65±2.54 & 96.52±3.11 & 86.60±3.05 & 98.07±1.91 & 98.10±1.03 \\
      & 8     & 96.91±2.80 & 98.25±1.28 & 84.66±6.90 & 97.67±1.53 & 97.56±2.33 & 88.30±4.34 & 98.70±1.02 & 98.34±1.14 \\
      & 16    & 98.96±0.39 & 99.23±0.46 & 90.32±2.84 & 98.67±0.71 & 99.24±0.44 & 92.61±3.27 & 99.46±0.31 & 99.09±0.67 \\
\bottomrule
\end{tabular}%
    }
    
\end{table*}%

\begin{table*}[h]
  \centering
  \caption{AUC across 10 datasets with different distance metrics.}
  \resizebox{0.95\linewidth}{!}{
\begin{tabular}{c|l|rrrrrr}
\toprule
\multirow{2}[4]{*}{dataset} & \multicolumn{1}{c|}{\multirow{2}[4]{*}{distance}} & \multicolumn{6}{c}{shot} \\
\cmidrule{3-8}      &       & 0     & 4     & 8     & 16    & 32    & 64 \\
\midrule
\multirow{3}[2]{*}{Adult} & Euclidean & \multicolumn{1}{l}{83.72±1.41} & \multicolumn{1}{l}{83.74±1.50} & \multicolumn{1}{l}{83.95±1.61} & \multicolumn{1}{l}{84.00±1.56} & \multicolumn{1}{l}{83.97±1.48} & \multicolumn{1}{l}{84.25±1.44} \\
      & Manhattan & \multicolumn{1}{l}{87.53±1.00} & \multicolumn{1}{l}{87.39±0.99} & \multicolumn{1}{l}{87.43±1.02} & \multicolumn{1}{l}{87.36±1.02} & \multicolumn{1}{l}{87.16±1.04} & \multicolumn{1}{l}{87.01±0.76} \\
      & Cosine & \multicolumn{1}{l}{83.84±1.38} & \multicolumn{1}{l}{83.85±1.51} & \multicolumn{1}{l}{84.04±1.57} & \multicolumn{1}{l}{83.99±1.48} & \multicolumn{1}{l}{83.83±1.42} & \multicolumn{1}{l}{84.02±1.41} \\
\midrule
\multirow{3}[2]{*}{Bank} & Euclidean & \multicolumn{1}{l}{82.93±1.90} & \multicolumn{1}{l}{82.53±1.73} & \multicolumn{1}{l}{82.71±1.95} & \multicolumn{1}{l}{83.13±1.60} & \multicolumn{1}{l}{83.86±1.24} & \multicolumn{1}{l}{84.52±0.64} \\
      & Manhattan & \multicolumn{1}{l}{80.22±1.55} & \multicolumn{1}{l}{81.25±1.67} & \multicolumn{1}{l}{81.82±1.85} & \multicolumn{1}{l}{82.73±1.38} & \multicolumn{1}{l}{83.88±1.07} & \multicolumn{1}{l}{84.81±0.88} \\
      & Cosine & \multicolumn{1}{l}{82.89±1.88} & \multicolumn{1}{l}{82.56±1.78} & \multicolumn{1}{l}{82.89±2.01} & \multicolumn{1}{l}{83.37±1.57} & \multicolumn{1}{l}{84.11±1.29} & \multicolumn{1}{l}{84.77±0.68} \\
\midrule
\multirow{3}[2]{*}{Blood} & Euclidean & \multicolumn{1}{l}{75.31±4.28} & \multicolumn{1}{l}{76.31±5.12} & \multicolumn{1}{l}{76.39±4.62} & \multicolumn{1}{l}{75.98±4.52} & \multicolumn{1}{l}{76.16±4.54} & \multicolumn{1}{l}{75.97±4.29} \\
      & Manhattan & \multicolumn{1}{l}{75.69±3.98} & \multicolumn{1}{l}{75.57±4.30} & \multicolumn{1}{l}{75.27±4.09} & \multicolumn{1}{l}{74.91±3.75} & \multicolumn{1}{l}{74.56±4.43} & \multicolumn{1}{l}{74.97±4.72} \\
      & Cosine & \multicolumn{1}{l}{74.58±4.22} & \multicolumn{1}{l}{76.01±4.79} & \multicolumn{1}{l}{75.48±4.47} & \multicolumn{1}{l}{76.04±4.30} & \multicolumn{1}{l}{76.59±4.56} & \multicolumn{1}{l}{76.23±4.11} \\
\midrule
\multirow{3}[2]{*}{Car} & Euclidean & \multicolumn{1}{l}{77.76±2.29} & \multicolumn{1}{l}{78.88±3.12} & \multicolumn{1}{l}{79.67±2.99} & \multicolumn{1}{l}{81.02±2.53} & \multicolumn{1}{l}{82.42±2.22} & \multicolumn{1}{l}{84.10±2.86} \\
      & Manhattan & \multicolumn{1}{l}{79.39±2.24} & \multicolumn{1}{l}{80.15±2.74} & \multicolumn{1}{l}{80.87±2.72} & \multicolumn{1}{l}{82.00±2.32} & \multicolumn{1}{l}{83.54±2.05} & \multicolumn{1}{l}{84.86±2.51} \\
      & Cosine & \multicolumn{1}{l}{77.64±2.27} & \multicolumn{1}{l}{78.78±2.95} & \multicolumn{1}{l}{79.68±2.91} & \multicolumn{1}{l}{81.27±2.62} & \multicolumn{1}{l}{82.95±2.34} & \multicolumn{1}{l}{85.26±3.02} \\
\midrule
\multirow{3}[2]{*}{Credit-g} & Euclidean & \multicolumn{1}{l}{59.64±3.15} & \multicolumn{1}{l}{61.43±4.19} & \multicolumn{1}{l}{63.26±3.73} & \multicolumn{1}{l}{65.22±3.46} & \multicolumn{1}{l}{69.03±2.52} & \multicolumn{1}{l}{72.60±2.53} \\
      & Manhattan & \multicolumn{1}{l}{62.62±3.01} & \multicolumn{1}{l}{63.86±3.25} & \multicolumn{1}{l}{65.52±3.09} & \multicolumn{1}{l}{67.12±2.79} & \multicolumn{1}{l}{70.15±2.78} & \multicolumn{1}{l}{72.66±2.49} \\
      & Cosine & \multicolumn{1}{l}{59.45±3.15} & \multicolumn{1}{l}{61.17±4.29} & \multicolumn{1}{l}{62.93±3.93} & \multicolumn{1}{l}{64.99±3.50} & \multicolumn{1}{l}{68.90±2.63} & \multicolumn{1}{l}{72.46±2.24} \\
\midrule
\multirow{3}[2]{*}{Cultivars} & Euclidean & \multicolumn{1}{l}{59.12±7.46} & \multicolumn{1}{l}{57.73±7.64} & \multicolumn{1}{l}{59.35±8.42} & \multicolumn{1}{l}{60.45±7.80} & \multicolumn{1}{l}{62.01±7.31} & \multicolumn{1}{l}{65.09±7.40} \\
      & Manhattan & \multicolumn{1}{l}{56.54±7.21} & \multicolumn{1}{l}{56.74±8.84} & \multicolumn{1}{l}{57.38±9.22} & \multicolumn{1}{l}{58.32±9.41} & \multicolumn{1}{l}{58.70±8.69} & \multicolumn{1}{l}{61.42±7.68} \\
      & Cosine & \multicolumn{1}{l}{59.12±7.46} & \multicolumn{1}{l}{57.67±7.75} & \multicolumn{1}{l}{59.41±8.31} & \multicolumn{1}{l}{60.72±7.60} & \multicolumn{1}{l}{62.94±7.20} & \multicolumn{1}{l}{65.16±7.77} \\
\midrule
\multirow{3}[2]{*}{Diabetes} & Euclidean & \multicolumn{1}{l}{80.32±2.03} & \multicolumn{1}{l}{80.45±2.29} & \multicolumn{1}{l}{80.48±2.26} & \multicolumn{1}{l}{80.46±2.38} & \multicolumn{1}{l}{81.02±2.71} & \multicolumn{1}{l}{81.50±2.20} \\
      & Manhattan & \multicolumn{1}{l}{79.79±2.40} & \multicolumn{1}{l}{80.04±2.56} & \multicolumn{1}{l}{80.32±2.56} & \multicolumn{1}{l}{80.14±2.50} & \multicolumn{1}{l}{80.81±2.53} & \multicolumn{1}{l}{80.82±2.17} \\
      & Cosine & \multicolumn{1}{l}{79.55±2.17} & \multicolumn{1}{l}{79.56±2.37} & \multicolumn{1}{l}{79.63±2.48} & \multicolumn{1}{l}{79.62±2.46} & \multicolumn{1}{l}{80.23±2.74} & \multicolumn{1}{l}{80.64±2.26} \\
\midrule
\multirow{3}[2]{*}{Heart} & Euclidean & \multicolumn{1}{l}{68.51±4.77} & \multicolumn{1}{l}{74.12±5.60} & \multicolumn{1}{l}{77.52±4.99} & \multicolumn{1}{l}{81.56±3.57} & \multicolumn{1}{l}{83.94±3.15} & \multicolumn{1}{l}{86.25±2.24} \\
      & Manhattan & \multicolumn{1}{l}{75.88±3.02} & \multicolumn{1}{l}{79.24±3.60} & \multicolumn{1}{l}{81.66±3.27} & \multicolumn{1}{l}{84.85±2.81} & \multicolumn{1}{l}{87.02±2.37} & \multicolumn{1}{l}{89.07±1.69} \\
      & Cosine & \multicolumn{1}{l}{69.43±4.53} & \multicolumn{1}{l}{74.50±5.48} & \multicolumn{1}{l}{77.64±5.08} & \multicolumn{1}{l}{81.46±3.49} & \multicolumn{1}{l}{83.77±3.20} & \multicolumn{1}{l}{86.18±2.21} \\
\midrule
\multirow{3}[2]{*}{Myocardial*} & Euclidean & \multicolumn{1}{l}{62.52±4.48} & \multicolumn{1}{l}{63.25±4.16} & \multicolumn{1}{l}{63.62±4.12} & \multicolumn{1}{l}{64.03±4.04} & \multicolumn{1}{l}{65.44±4.38} & \multicolumn{1}{l}{65.75±4.34} \\
      & Manhattan & \multicolumn{1}{l}{63.52±3.99} & \multicolumn{1}{l}{64.09±3.60} & \multicolumn{1}{l}{64.07±3.56} & \multicolumn{1}{l}{64.56±3.82} & \multicolumn{1}{l}{65.66±3.95} & \multicolumn{1}{l}{65.91±4.03} \\
      & Cosine & \multicolumn{1}{l}{62.29±4.51} & \multicolumn{1}{l}{63.08±4.14} & \multicolumn{1}{l}{63.16±4.29} & \multicolumn{1}{l}{63.77±4.54} & \multicolumn{1}{l}{65.03±4.59} & \multicolumn{1}{l}{65.50±4.66} \\
\midrule
\multirow{3}[2]{*}{NHANES} & Euclidean & \multicolumn{1}{l}{97.48±1.20} & \multicolumn{1}{l}{98.10±1.03} & \multicolumn{1}{l}{98.34±1.14} & \multicolumn{1}{l}{99.09±0.67} & \multicolumn{1}{l}{99.42±0.43} & \multicolumn{1}{l}{99.65±0.20} \\
      & Manhattan & \multicolumn{1}{l}{95.43±1.57} & \multicolumn{1}{l}{96.22±1.41} & \multicolumn{1}{l}{96.66±1.64} & \multicolumn{1}{l}{97.95±1.06} & \multicolumn{1}{l}{98.67±0.81} & \multicolumn{1}{l}{99.33±0.41} \\
      & Cosine & \multicolumn{1}{l}{97.55±1.12} & \multicolumn{1}{l}{98.12±0.95} & \multicolumn{1}{l}{98.43±0.93} & \multicolumn{1}{l}{99.10±0.48} & \multicolumn{1}{l}{99.31±0.32} & \multicolumn{1}{l}{99.42±0.23} \\
\midrule
\multirow{3}[2]{*}{Average} & Euclidean & 74.73  & 75.65  & 76.53  & 77.49  & 78.73  & 79.97  \\
      & Manhattan & 75.66  & 76.46  & 77.10  & 77.99  & 79.02  & 80.09  \\
      & Cosine & 74.63  & 75.53  & 76.33  & 77.43  & 78.77  & 79.96  \\
\bottomrule
\end{tabular}%

    }
  \label{tab: distance metrics}%
\end{table*}%

\label{appendix: Data Augmentation}
\begin{table*}[h]
  \centering
  \caption{Applying Data Augmentation to LogReg,KNN, and MLP.}
  \resizebox{\linewidth}{!}{
\begin{tabular}{c|c|cc|cc|cc|cc|cc|cc}
\toprule
\multirow{3}[6]{*}{Dataset} & \multirow{3}[6]{*}{Model} & \multicolumn{12}{c}{shot} \\
\cmidrule{3-14}      &       & \multicolumn{2}{c|}{0} & \multicolumn{2}{c|}{4} & \multicolumn{2}{c|}{8} & \multicolumn{2}{c|}{16} & \multicolumn{2}{c|}{32} & \multicolumn{2}{c}{64} \\
\cmidrule{3-14}      &       & w/o Aug & w/ Aug & w/o Aug & w/ Aug & w/o Aug & w/ Aug & w/o Aug & w/ Aug & w/o Aug & w/ Aug & w/o Aug & w/ Aug \\
\midrule
\multirow{3}[2]{*}{Adult} & LogReg & N/A   & \textbf{76.52} & 65.52 & \textbf{76.83} & 71.9  & \textbf{80.98} & 78.27 & \textbf{82.6} & 81.82 & \textbf{83.73} & 84.54 & \textbf{85.21} \\
      & KNN   & N/A   & \textbf{82.79} & 61.39 & \textbf{83.41} & 72.43 & \textbf{83.84} & 78.25 & \textbf{84.14} & 81.59 & \textbf{84.52} & 84.14 & \textbf{85.09} \\
      & MLP   & N/A   & \textbf{71.63} & 55.16 & \textbf{80.2} & 71.24 & \textbf{81.45} & 78.01 & \textbf{85.4} & 80.91 & \textbf{85.95} & 85.39 & \textbf{85.43} \\
\midrule
\multirow{3}[2]{*}{Bank} & LogReg & N/A   & \textbf{67.83} & 59.29 & \textbf{76.07} & 66.46 & \textbf{78.2} & 74.15 & \textbf{81.51} & 78.25 & \textbf{81.01} & 81.61 & \textbf{84.32} \\
      & KNN   & N/A   & \textbf{75.81} & 57.96 & \textbf{76.87} & 63.13 & \textbf{77.17} & 69.38 & \textbf{78.34} & 73.43 & \textbf{80.35} & 77.8  & \textbf{82.44} \\
      & MLP   & N/A   & \textbf{65.6} & 56.87 & \textbf{72.18} & 61.86 & \textbf{73.63} & 66.8  & \textbf{78.22} & 71.99 & \textbf{80.92} & 79.35 & \textbf{82.67} \\
\midrule
\multirow{3}[2]{*}{Blood} & LogReg & N/A   & \textbf{74.02} & 58.02 & \textbf{73.53} & 57.2  & \textbf{75.4} & 65.41 & \textbf{76.76} & 72.3  & \textbf{76.12} & 74.86 & \textbf{76.76} \\
      & KNN   & N/A   & \textbf{73.18} & 53.33 & \textbf{74.41} & 57.73 & \textbf{74.7} & 64.68 & \textbf{74.79} & 68.55 & \textbf{74.75} & 72.46 & \textbf{75.04} \\
      & MLP   & N/A   & \textbf{69.32} & 54.9  & \textbf{68.51} & 63.4  & \textbf{67.49} & 65.92 & \textbf{73.94} & 62.82 & \textbf{73.22} & 72.72 & \textbf{73.19} \\
\midrule
\multirow{3}[2]{*}{Car} & LogReg & N/A   & \textbf{68.78} & 65.14 & \textbf{72.84} & 65.05 & \textbf{76.79} & 76.33 & \textbf{79.7} & \textbf{84.95} & 82.07 & \textbf{91.69} & 87.67 \\
      & KNN   & N/A   & \textbf{72.87} & 60.14 & \textbf{73.93} & 65.02 & \textbf{74.37} & 72.23 & \textbf{75.59} & 77.5  & \textbf{78.88} & 81.32 & \textbf{83.18} \\
      & MLP   & N/A   & \textbf{76.17} & 52.85 & \textbf{76.25} & 62.9  & \textbf{77.84} & 75.34 & \textbf{80.69} & 83.56 & \textbf{83.65} & \textbf{87.86} & 87.4 \\
\midrule
\multirow{3}[2]{*}{Credit-g} & LogReg & N/A   & \textbf{53.91} & 54.01 & \textbf{57.92} & 58.15 & \textbf{62.64} & 58.62 & \textbf{64.48} & 64.16 & \textbf{68.82} & 68.51 & \textbf{71.27} \\
      & KNN   & N/A   & \textbf{59.23} & 53.33 & \textbf{58.48} & 54.26 & \textbf{58.92} & 56.89 & \textbf{61.87} & 61.06 & \textbf{65.96} & 66.06 & \textbf{68.51} \\
      & MLP   & N/A   & \textbf{56.17} & 51.27 & \textbf{60.21} & 53.6  & \textbf{63.1} & 55.39 & \textbf{62.51} & 60.5  & \textbf{66.96} & 65.96 & \textbf{69.34} \\
\midrule
\multirow{3}[2]{*}{Cultivars} & LogReg & N/A   & \textbf{60.34} & 44.98 & \textbf{51.98} & 50.2  & \textbf{55.99} & 48.48 & \textbf{59.3} & 53.15 & \textbf{62.56} & 63.7  & \textbf{69.55} \\
      & KNN   & N/A   & \textbf{53.69} & 45.84 & \textbf{58.65} & 47.47 & \textbf{57.55} & 47.62 & \textbf{54.81} & 50.54 & \textbf{55.19} & 51.01 & \textbf{54} \\
      & MLP   & N/A   & \textbf{60.81} & 43.95 & \textbf{56.69} & 47.64 & \textbf{58.61} & 48.86 & \textbf{62.7} & 57.2  & \textbf{66.37} & 67.51 & \textbf{73.79} \\
\midrule
\multirow{3}[2]{*}{Diabetes} & LogReg & N/A   & \textbf{75.53} & 58.74 & \textbf{76.69} & 70.79 & \textbf{75.14} & 66.34 & \textbf{76.01} & \textbf{77.44} & 77.39 & 78.72 & \textbf{79.78} \\
      & KNN   & N/A   & \textbf{71.74} & 59.48 & \textbf{71.7} & 63.7  & \textbf{72.23} & 68.3  & \textbf{73.13} & 73.96 & \textbf{75.5} & \textbf{77.59} & 76.22 \\
      & MLP   & N/A   & \textbf{69.17} & 58.53 & \textbf{72.71} & 63.73 & \textbf{66.8} & 66.61 & \textbf{73.76} & 74.39 & \textbf{76.11} & \textbf{77} & 75.43 \\
\midrule
\multirow{3}[2]{*}{Heart} & LogReg & N/A   & \textbf{56.65} & 62.25 & \textbf{72.76} & 77.44 & \textbf{78.97} & 84.99 & \textbf{85.82} & \textbf{88.41} & 87.8  & \textbf{89.62} & 89.41 \\
      & KNN   & N/A   & \textbf{62.57} & 63.07 & \textbf{67.71} & 74.37 & \textbf{71.74} & \textbf{84.53} & 78.11 & \textbf{87.96} & 83.71 & \textbf{89} & 88.32 \\
      & MLP   & N/A   & \textbf{56.62} & 56.24 & \textbf{69.44} & 79.07 & \textbf{78.03} & 87.12 & \textbf{87.36} & 87.8  & \textbf{88.54} & \textbf{89.04} & 89.38 \\
\midrule
\multirow{3}[2]{*}{Myocardial} & LogReg & N/A   & \textbf{60.15} & 54.88 & \textbf{62.57} & 56.36 & \textbf{62.38} & 54.77 & \textbf{63.5} & 63.03 & \textbf{65.34} & 64.49 & \textbf{68.76} \\
      & KNN   & N/A   & \textbf{50} & 53.28 & \textbf{54.44} & 55.92 & \textbf{57.61} & 54.49 & \textbf{58.36} & 58.88 & \textbf{59.13} & 60.04 & \textbf{60.46} \\
      & MLP   & N/A   & \textbf{54.25} & 53.87 & \textbf{61.99} & 55.25 & \textbf{62.67} & 55.21 & \textbf{62.98} & 57.21 & \textbf{65.17} & 58.81 & \textbf{65} \\
\midrule
\multirow{3}[2]{*}{NHANES} & LogReg & N/A   & \textbf{98.8} & 77.56 & \textbf{98.78} & 88.34 & \textbf{97.17} & 97.12 & \textbf{98.9} & 99.3  & \textbf{99.58} & 99.49 & \textbf{99.86} \\
      & KNN   & N/A   & \textbf{79.17} & 64.9  & \textbf{80.62} & 75.38 & \textbf{83.01} & 86.34 & \textbf{87.77} & 92.18 & \textbf{92.34} & 94.79 & \textbf{95.98} \\
      & MLP   & N/A   & \textbf{74.25} & 59.58 & \textbf{80.09} & 76.43 & \textbf{84.91} & 86.34 & \textbf{93.21} & 91.88 & \textbf{95.01} & 95.32 & \textbf{98.84} \\
\bottomrule
\end{tabular}%

    }
  \label{tab:Data Augmentation}%
\end{table*}%
\clearpage
\end{document}